\documentclass[10pt,twocolumn,letterpaper]{article}

\usepackage[pagenumbers]{cvpr} % To force page numbers, \eg for an arXiv version

\usepackage{graphicx}
\usepackage{amsmath}
\usepackage{amssymb}
\usepackage{booktabs}
\usepackage{mathtools}
\usepackage{multirow}
\usepackage{tablefootnote}
\usepackage[dvipsnames]{xcolor}
\newcommand{\tabincell}[2]{\begin{tabular}{@{}#1@{}}#2\end{tabular}} 
\usepackage{color}
\usepackage{algorithm, algpseudocode}
\usepackage{bm}
\newtheorem{definition}{Definition}[section]

\definecolor{myred}{RGB}{205 38 38}
\usepackage[pagebackref,breaklinks,colorlinks]{hyperref}

\usepackage[capitalize]{cleveref}
\crefname{section}{Sec.}{Secs.}
\Crefname{section}{Section}{Sections}
\Crefname{table}{Table}{Tables}
\crefname{table}{Tab.}{Tabs.}

\def\cvprPaperID{6628} % *** Enter the CVPR Paper ID here
\def\confName{CVPR}
\def\confYear{2023}

\begin{document}

\title{Towards Effective Adversarial Textured 3D Meshes on Physical Face Recognition}

\renewcommand\footnotemark{}
\author{Xiao Yang$^{1}$, Chang Liu$^{2}$, Longlong Xu$^{1}$, Yikai Wang$^{1}$, Yinpeng Dong$^{1,3\dagger}$, \\Ning Chen$^{1}$, Hang Su$^{1,4}$, Jun Zhu$^{1,3,4\dagger}$\thanks{$^\dagger$Corresponding authors.} \\
$^{1}$ Dept. of Comp. Sci. and Tech., Institute for AI, BNRist Center, Tsinghua-Bosch Joint ML Center,\\
THBI Lab, Tsinghua-China Mobile
Communications Group Co., Ltd. Joint Institute,
Tsinghua University\\ 
$^{2}$ Peking University \hspace{2ex} $^{3}$ RealAI \hspace{2ex}
$^{4}$ Zhongguancun Laboratory\\
\tt\small{\{yangxiao19, xu-ll18\}@mails.tsinghua.edu.cn} \hspace{2ex}\tt\small{chang.liu@stu.pku.edu.cn} \hspace{2ex} \tt\small{yikaiw@outlook.com}  \\ \tt\small{\{dongyinpeng, ningchen, suhangss, dcszj\}@tsinghua.edu.cn} \\
}
\maketitle
%As face recognition being becoming a prevailing authentication solution in numerous biometric applications, physical adversarial attacks raise concerns about the security of the deployed face recognition systems, especially in safety-critical cases.
%%%%%%%%% ABSTRACT
\begin{abstract}
Face recognition is a prevailing authentication solution in numerous biometric applications. Physical adversarial attacks, as an important surrogate, can identify the weaknesses of face recognition systems and evaluate their robustness before deployed. However, most existing physical attacks are either detectable readily or ineffective against commercial recognition systems. The goal of this work is to develop a more reliable technique that can carry out an end-to-end evaluation of adversarial robustness for commercial systems. It requires that this technique can simultaneously deceive black-box recognition models and evade defensive mechanisms. To fulfill this, we design adversarial textured 3D meshes (\textbf{AT3D}) with an elaborate topology on a human face, which can be 3D-printed and pasted on the attacker's face to evade the defenses. However, the mesh-based optimization regime calculates gradients in high-dimensional mesh space, and can be trapped into local optima with unsatisfactory transferability. To deviate from the mesh-based space, we propose to perturb the low-dimensional coefficient space based on 3D Morphable Model, which significantly improves black-box transferability meanwhile enjoying faster search efficiency and better visual quality. Extensive experiments in digital and physical scenarios show that our method effectively explores the security vulnerabilities of multiple popular commercial services, including \textbf{three} recognition APIs, \textbf{four} anti-spoofing APIs, \textbf{two} prevailing mobile phones and \textbf{two} automated access control systems.
% \hangx{we may emphasize that our method is effective on some popular commercial services and identify their potential risk, which would be more attractive}

\end{abstract}

%%%%%%%%% BODY TEXT
%\vspace{-0.2cm}
\section{Introduction}
\label{sec:intro}
%\vspace{-0.1cm}

Face recognition has become a prevailing authentication solution in biometric applications, ranging from financial payment to automated surveillance systems. Despite its blooming development~\cite{deng2019arcface,wang2018cosface,schroff2015facenet}, recent research in adversarial machine learning has revealed that face recognition models based on deep neural networks are highly vulnerable to adversarial examples~\cite{goodfellow2014explaining,yang2021model}, leading to serious consequences or security problems in real-world applications.

%Automatic face recognition systems are capable of identifying or verifying a person from visual face images.
%Empowered by the excellent performance of deep neural networks (DNNs)~\cite{simonyan2014very,szegedy2015going,he2015deep},Automatic face recognition systems are capable of identifying or verifying a person from visual face images. 

Due to the imperative need of evaluating model robustness~\cite{yang2020delving,tong2021facesec}, extensive attempts have been devoted to adversarial attacks on face recognition models. Adversarial attacks in the digital world~\cite{sharif2016accessorize,yang2020delving,dong2019efficient,yang2021towards} are characterized by adding minimal perturbations to face images in the \emph{digital} space, aiming to evade being recognized or to impersonate another identity. Since an adversary usually cannot access the digital input of practical systems, physical adversarial examples wearable for real human faces are more feasible for evaluating their adversarial robustness. Some studies have shown the success of physical attacks against popular recognition models by adopting different attack types, such as eyeglass frames \cite{sharif2016accessorize,sharif2017adversarial}, hats~\cite{komkov2021advhat} and stickers~\cite{shen2021effective}. 

%The crafted adversarial examples in black-box manner~\cite{dong2019efficient} also successfully mislead commercial face recognition APIs. 

\begin{figure}[t]
\begin{center}
\includegraphics[width=0.99\linewidth]{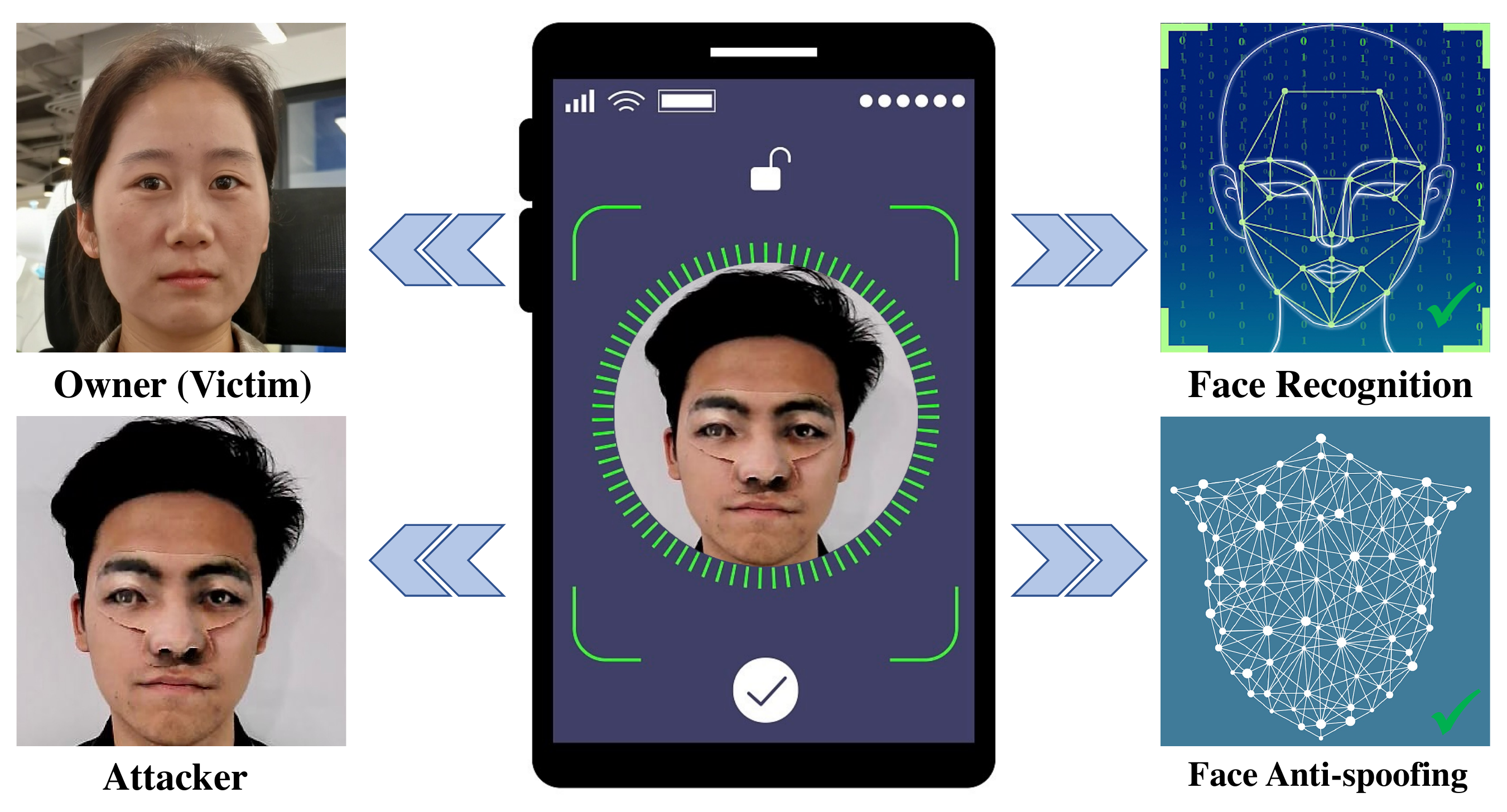}
\end{center}
\vspace{-4ex}
\caption{Demonstration of physical black-box attacks for unlocking one prevailing mobile phone. The attacker wearing the 3D-printed adversarial mesh can successfully mislead the face recognition model to be recognized as the victim, meanwhile evading face anti-spoofing. More results are shown in Sec.~\ref{sec:exps}. }
\label{fig:intro}
\vspace{-3ex}
\end{figure}

%%%%%%%%%%%%%%%%%%%%%%%% related %%%%%%%%%%%%%
\begin{table*}[t]
\small 
% \footnotesize
\setlength{\tabcolsep}{2.5pt}
\begin{center}
\begin{tabular}{c|cccccccc}
\hline
& Frames~\cite{sharif2016accessorize} & AdvHat~\cite{komkov2021advhat} & FaceAdv~\cite{shen2021effective} & PadvFace~\cite{zheng2021robust} & AdvMask~\cite{zolfi2021adversarial} &  Face3DAdv~\cite{yang2022controllable} & RHDE~\cite{wei2022adversarial} &Ours \\
\hline
\tabincell{c}{3D attack types} & No & \emph{Partially} & \emph{Partially}& No &\textbf{Yes}& \textbf{Yes} & \emph{Partially} & \textbf{Yes} \\

\tabincell{c}{Commercial recognition} & \textbf{Yes} &  No & No & No & No & No & \textbf{Yes}& \textbf{Yes}\\
\tabincell{c}{Commercial defenses} &  No & No & No & No &  No & \textbf{Yes}& No &\textbf{Yes}\\
\tabincell{c}{Number of physical tests} &10 & 3 & 10 & 10 & 30 & 10 & 3 &50\\
\hline
\end{tabular}
\end{center}
\vspace{-4ex}
\caption{A comparison among different methods regarding whether using 3D attack types, commercial face recognition models, commercial defenses, and the number of physical  evaluation. \emph{Partially} indicates that this method involved some geometric transformations to make 2D patch approximately approach the realistic 3D patch.}
\vspace{-3ex}
\label{tab:related-work}
\end{table*}
%%%%%%%%%%%%%%%%%%%%%%%%%%%%%%%%%

In spite of the remarkable progress, it is challenging to launch \emph{practical} and \emph{effective} physical attack methods on automatic face recognition systems. First, the defensive mechanism~\cite{jia2021dual,zhang2020casia,yu2020searching,yang2022towards,yang2019face} on face recognition, \ie, face anti-spoofing, has achieved impressive performance among the academic and industry communities. Some popular defenses~\cite{zhang2019dataset,liu2021casia,wang2020deep} have injected more sensors (such as depth, multi-spectral and infrared cameras) to provide more effective defenses. However, most of the physical attacks have not evaluated the passing rates against practical defensive mechanisms, as reported in Table.~\ref{tab:related-work}. Second, these methods cannot perform satisfactorily for impersonation attacks against diverse commercial black-box recognition models due to the limited black-box transferability. The goal of this work is to develop \emph{practical} and \emph{effective} physical adversarial attacks that can simultaneously deceive black-box recognition models and evade defensive mechanisms in commercial face recognition systems, \eg, unlocking mobile phones, as demonstrated in Fig.~\ref{fig:intro}.

% steadily pass the automatic FR system with black-box recognition and defensive mechanisms.

% The goal of this work is to develop a method that can % evading defensive mechanism in face recognition systems.impersonating one’s identity, meanwhile 

\textbf{Evading the defensive mechanisms.} Recent research has found that high-fidelity 3D masks~\cite{liu20163d,liu20213d} can better fool the prevailing face anti-spoofing methods by 3D printing techniques. It becomes an appealing and feasible way to apply a 3D adversarial mask for evading defensive mechanisms in face recognition systems. To achieve this goal, we first design adversarial textured 3D meshes (\textbf{AT3D}) with an elaborate topology on a human face, which can be usable by standard graphics software such as Blender~\cite{flavell2011beginning} and Maya~\cite{Maya}. As a primary 3D representation, textured meshes can be immediately 3D-printed and pasted on real faces for physical adversarial attacks, which have geometric details, complex topology and high-quality textures. Experimentally, AT3D can be more conducive to steadily passing commercial face anti-spoofing services, such as FaceID and Tencent anti-spoofing APIs, two mobile phones and two access control systems with \emph{multiple sensors}.
%  After introducing differentiable rendering~\cite{ravi2020accelerating}, we can directly optimize the adversarial mesh only using 2D victim images, which are more available than explicit 3D mesh. \hangx{what points? may have some background} \hangx{why lower dimension is better?}

\textbf{Misleading the black-box recognition models.} 
The typical 3D mesh attacks~\cite{xiao2019meshadv,zhang20213d,miaoisometric} proposed to optimize adversarial examples in mesh representation space. Thus, high complexity is virtually inevitable for calculating gradients in such high-dimensional search space due to the thousands of triangle faces on each human face.
The procedures are also costly and probably trapped into overfitting~\cite{liu2022imperceptible} with unsatisfactory transferability. Therefore, we aim to perform the optimization trajectory in a low-dimensional manifold as a regularization aiming for escaping from overfitting.  The low-dimensional manifold should possess a sufficient capacity that encodes any 3D face in this low-dimensional feature space, thus successfully achieving the white-box adversarial attack against a substitute model. A principled way of spanning such a subspace is considered by leveraging 3D Morphable Model (3DMM)~\cite{tuan2017regressing} that effectively achieves dimensionality reduction of any high-dimensional mesh data. Based on this, we are capable of generating an adversarial mesh by perturbing the low-dimensional coefficients of 3DMM, making it constrained on the data manifold of realistic 3D faces. Therefore, the crafted mesh can obtain a strong semantic feature of a 3D face, 
which can achieve well-generalizing performance among the white-box and black-box models due to knowledgable semantic pattern characteristics~\cite{xiao2021improving,yang2021boosting,yang2020design}. In addition, low-dimensional optimization can also avoid self-intersection and flying vertices problems in mesh-based optimization~\cite{zhang20213d}, resulting in better visual appearance.

% which  can relieve the overfitting problem since the predictions of white-box and black-box model are 
% which significantly improves black-box transferability. 
% \hangx{better more explicit. why low dimension is useful for transferability? why use 3DMM as for dimension reduction? any nontrivial challenges to be addressed? not just adopt some well studied method.}

% Overall, the adversarial textured 3D meshes not only significantly achieve high success rates of impersonation attacks against multiple commercial black-box recognition models but also steadily evade commercial defensive mechanisms. 
Experimentally, we have effectively explored the security vulnerabilities of multiple popular commercial services, including 1) recognition APIs---Amazon, Face++, and Tencent;  
2) anti-spoofing APIs---FaceID, SenseID, Tencent, and Aliyun; 3) \textbf{two} prevailing mobile phones and \textbf{two} automated access control systems that incorporate multiple sensors. Our main contributions can be summarized as:
\begin{itemize}
    \item We propose effective and practical adversarial textured 3D meshes with elaborate topology and effective optimization, which can simultaneously evade black-box recognition models and defensive mechanisms.
    \item Extensive physical experiments demonstrate that our method can consistently mislead multiple commercial systems, including unlocking prevailing mobile phones and automated access control systems.
    
    \item We present a reliable technique to evaluate the robustness of face recognition systems, which can be further leveraged as an effective data augmentation strategy to improve defensive ability.
    % promote the development of more robust models.
    % \hangx{maybe not a technical contribution}
\end{itemize}

% can be directly consumed by 3D rendering engines

\begin{figure*}[t]
\begin{center}
\includegraphics[width=0.99\linewidth]{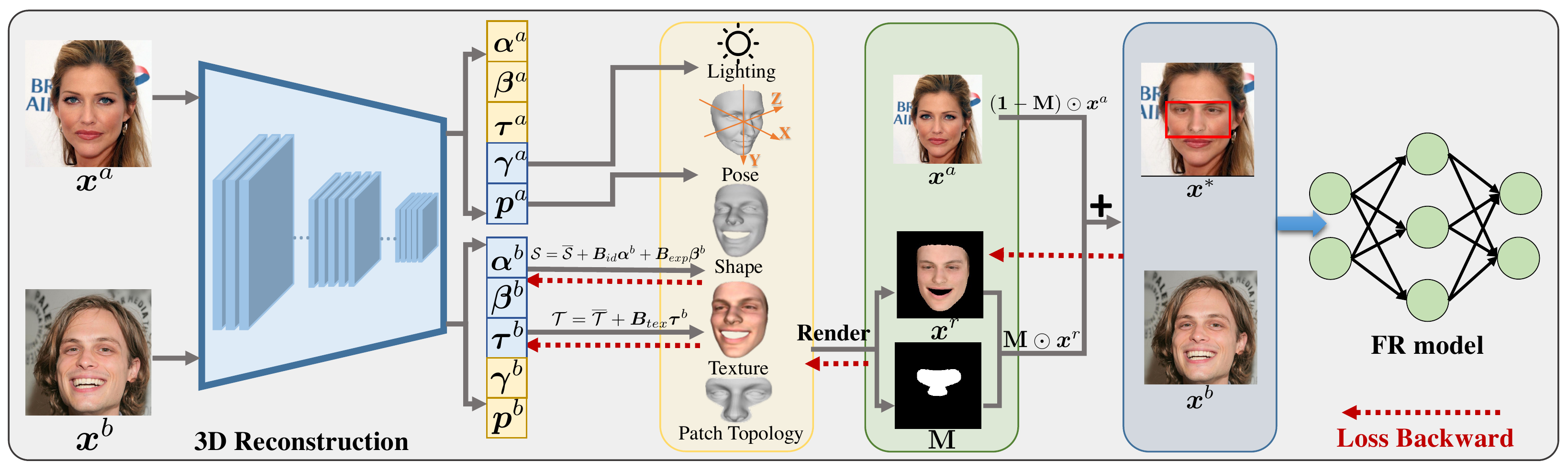}
\end{center}
\vspace{-4ex}
\caption{An overview of crafting adversarial textured 3D meshes in the low-dimensional manifold. The 3D reconstruction model first regresses the coefficients of 3DMM, \ie, $\{\bm{\alpha}, \bm{\beta}, \bm{\tau}, \bm{\gamma}, \bm{p}\}$. Thus the shape and texture can be calculated by using the calculated coefficients. After introducing the elaborate local topology, the adversarial generation can be restricted to a specifically designed region. After rendering, we can obtain a rendered image $\bm{x}^r$ and a calculated 2D binary matrix $\mathbf{M}$. Since the whole pipeline including the rendering procedure is differentiable, the adversarial mesh can be iteratively updated by backpropagation on the low-dimensional coefficient space of 3DMM.}
\label{fig:frame}
\vspace{-3ex}
\end{figure*}

\vspace{-0.2cm}
\section{Related Work}
\vspace{-0.1cm}
In this section, we review related work about physical adversarial attacks on face recognition, and present a detailed comparison between the different methods in Table~\ref{tab:related-work}.

\textbf{2D physical adversarial attacks on face recognition.} Several early works have been developed to craft adversarial patches in the physical world~\cite{sharif2016accessorize,sharif2017adversarial} against face recognition systems. By pasting a carefully crafted 2D patch to the face, some research~\cite{pautov2019adversarial,komkov2021advhat} has shown effective physical attacks against state-of-the-art face recognition algorithms. AdvHat~\cite{komkov2021advhat} adopted the mask type of Hat to achieve an impersonation attack. However, the aforementioned 2D methods are required to be placed on relatively flat regions, limiting practicality when fitting the patch to the real 3D face.

\textbf{3D physical adversarial attacks on face recognition.} Some studies~\cite{yang2022controllable,zolfi2021adversarial} have exploited simple geometric transformations of the patch for approximatively achieving realistic 3D fitting procedures, \eg, parabolic transformation~\cite{komkov2021advhat,wei2022adversarial}. Furthermore, 3D affine transformation can be applied to the patch for simulating the corresponding pitch rotation. Besides, some 3D patches~\cite{zolfi2021adversarial,yang2022controllable} can be naturally stitched onto the face to make the adversarial patch realistic by fully leveraging the recent advances in 3D face modeling. However, these techniques are only either perceptually satisfactory or ineffective against black-box face recognition systems. As a comparison, ours can simultaneously deceive black-box recognition models and evade defensive mechanisms in commercial face recognition systems.

\vspace{-0.2cm}
\section{Method}
\vspace{-0.1cm}
\label{sec:method}
We first propose adversarial textured 3D meshes (\textbf{AT3D}) that can bypass general defensive mechanisms in Sec.~\ref{sec:formulation}. Afterwards, we propose a low-dimensional optimization to boost the transferability of the attack methods in Sec.~\ref{sec:iat3d}. An overview of our proposed method is provided in Fig.~\ref{fig:frame}.

\vspace{-0.1cm}
\subsection{Preliminary}
\label{sec:pre}
\vspace{-0.1cm}
Face recognition consists of two sub-tasks~\cite{goodfellow2014explaining}, \ie, face verification and face identification. The former aims to distinguish whether a pair of facial images belong to the same identity, while the latter directly predicts the identity of the facial image. We mainly study face verification in this paper, since the attacks can be easily extended to face identification. Denote $f(\bm{x}): \mathcal{X}\rightarrow\mathbb{R}^d$ as a face recognition model that outputs a feature representation in $\mathbb{R}^d$. In face verification, the similarity~\cite{deng2018arcface,wang2018cosface} between a pair of images $\{\bm{x}^{a}, \bm{x}^{b}\} \subset \mathcal{X}$ can be commonly calculated as 
\begin{equation}
\label{eq:Sf}
    {J}_f(\bm{x}^{a}, \bm{x}^{b}) = \frac{<f(\bm{x}^{a}), f(\bm{x}^{b})>}{\| f(\bm{x}^{a})\|\cdot \| f(\bm{x}^{b})\|},
\end{equation}
where $<\bm{\cdot}, \bm{\cdot}>$ is the inner product of the vectors. ${J}_f$ refers to cosine similarity between feature representations of $\bm{x}^{a}$ and $\bm{x}^{b}$ ranging from 0 to 1. Then the prediction of face verification can be formulated as
\begin{equation}
\label{eq:classifier}
    \mathcal{C}(\bm{x}^{a}, \bm{x}^{b}) = \mathbb{I}(J_f(\bm{x}^{a}, \bm{x}^{b})>\delta),
\end{equation}
where $\mathbb{I}$ is the indicator function, and $\delta$ is a threshold. When $\mathcal{C}(\bm{x}^{a}, \bm{x}^{b})$ equals to 1, the two images are regarded as the same identity, otherwise different identities.

% Given face images $\bm{x}^{a}$ and $\bm{x}^{b}$, we aim to generate an adversarial image $\bm{x}^*$ by adding a perturbation to $\bm{x}^{a}$ to fool the face recognition model. 
% \junz{why consider these two attacks? give reasons, e.g., are they the only two or the most practical/challenging ones?}
We focus on two general types of attacks in terms of \textbf{dodging} and \textbf{impersonation} with different goals.  In a dodging attack, an attacker attempts to fool a face recognition system by making one face misidentified, generally bypassing a face recognition system in surveillance. Formally,  the attacker aims to modify $\bm{x}$ to craft an adversarial image $\bm{x}^{*}$ to make $\mathcal{C} (\bm{x}^{*}, \bm{x}^{b}) = 0$ while $\mathcal{C} (\bm{x}^{a}, \bm{x}^{b}) = 1$. In contrast, an impersonation attack intends to disguise the attacker as another target identity. The generated adversarial image $\bm{x}^{*}$ will be recognized as the target identity of $\bm{x}^{b}$ that makes $\mathcal{C} (\bm{x}^{*}, \bm{x}^{b}) = 1$ while $\mathcal{C} (\bm{x}^{a}, \bm{x}^{b}) = 0$.

\subsection{Problem Formulation}
\label{sec:formulation}
% \hangx{
% I think you may rephrase this section, to identify your novelty and key contribution;
% \begin{itemize}
%     \item what problem we are solve and what is the benefit: 3d face adversarial attacks
%     \item why use mesh-wise representation 
%     \item formulate your problem 
%     \item the nontrivial challenges to solve the problem which can motivate your dimension reduction technique, etc. 
% \end{itemize}
% }

For the 3D adversarial attack task, we aim to develop an effective approach that can simultaneously deceive black-box recognition models and evade defenses in physical face recognition systems. Different from the existing 3D attacks in point clouds~\cite{zhang20213d}, we propose to craft an adversarial textured 3D mesh with any topology to avoid large errors by reconstruction procedure in point clouds~\cite{varley2017shape}. In addition, textured meshes can fully leverage 3D printing techniques for physically realizable adversarial attacks.

Specifically, we denote the mesh representation of a full face as $\mathcal{M} =(\mathcal{S},\mathcal{T},\mathcal{F})$,  where $\mathcal{S}\in\mathbb{R}^{n\times 3}$ is the $xyz$ coordinates of $n$ vertices, $\mathcal{T}\in\mathbb{R}^{n\times 3}$ is the $rgb$ value of vertices, and $\mathcal{F}\in\mathbb{Z}^{m\times 3}$ is the set of $m$ triangle faces which encodes each triangle with the indices of vertices. In addition, we are capable of studying 3D adversarial patch $\mathcal{M}'$ that is restricted to a specifically designed spacial region, which can be generated by modifying the original mesh topology $\mathcal{F}$. The topology $\mathcal{F}'$ of the 3D adversarial patch stems from a subset of the original $\mathcal{F}$ by erasing the triangle faces outside the patch, thus denoted as $\mathcal{M}'=(\mathcal{S},\mathcal{T},\mathcal{F}')$.

% Based on differential rendering in Eq.~\eqref{eq:rendering},

In this paper, we focus on crafting the \textbf{A}dversarial \textbf{T}extured \textbf{3D} meshes (\textbf{AT3D}) by modifying the vertex positions and colors.
Formally, an adversarial mesh can be denoted as $\mathcal{M}^{*}=(\mathcal{S}^{*}, \mathcal{T}^{*}, \mathcal{F}')$ by directly optimizing $\mathcal{S}$ and $\mathcal{T}$. Since 2D victim images are usually more available than the corresponding explicit 3D mesh, we consider converting 3D mesh representation into 2D images for optimization by introducing differentiable neural rendering~\cite{ravi2020accelerating}. Therefore, the attack objective function of crafting adversarial examples can be formulated as 
% \begin{equation}
% \small
\begin{gather}
    \min_{\mathcal{S}^{*}, \mathcal{T}^{*}}\; \mathcal{L}_{f}(\bm{x}^{*}, \bm{x}^{b}), \;
    \text{where }
    \bm{x}^{*} = \mathbf{M} \odot \bm{x}^{r} + (\bm{1} - \mathbf{M})\odot\bm{x}^{a}, \nonumber\\
     \bm{x}^{r},\mathbf{M}=\mathrm{R}(\mathcal{S}^{*},\mathcal{T}^{*}, \mathcal{F'}, \bm{\rho}),\label{eq:object}
\end{gather}
% \end{equation}
where $\odot$ is the element-wise multiplication operation and $\mathrm{R}$ is the rendering function. Given the rendering parameters $\bm{\rho}$ that contain camera position and illumination intensity, we can obtain 1) a rendered image $\bm{x}^r$ by rendering the mesh $\mathcal{M}'$ onto a black background; 2) a calculated 2D binary matrix $\mathbf{M}$ that takes $0$ if the pixel value derives from the background, and $1$ otherwise. In this paper, we adopt the attack loss $\mathcal{L}_{f} = J_{f}$ for a dodging attack and $\mathcal{L}_{f} = -{J}_{f}$ for an impersonation attack. By optimizing problem~(\ref{eq:object}) given a 2D face image $\bm{x}^{a}$, we can obtain the adversarial mesh $\mathcal{M}^{*}=(\mathcal{S}^{*}, \mathcal{T}^{*}, \mathcal{F}')$. 
% \hangx{more formal analysis. what are the nontrivial challenges? why we need mesh-wise representation? why we need low dimension optimization? }

\begin{figure}[t]
\begin{center}
\includegraphics[width=0.9\linewidth]{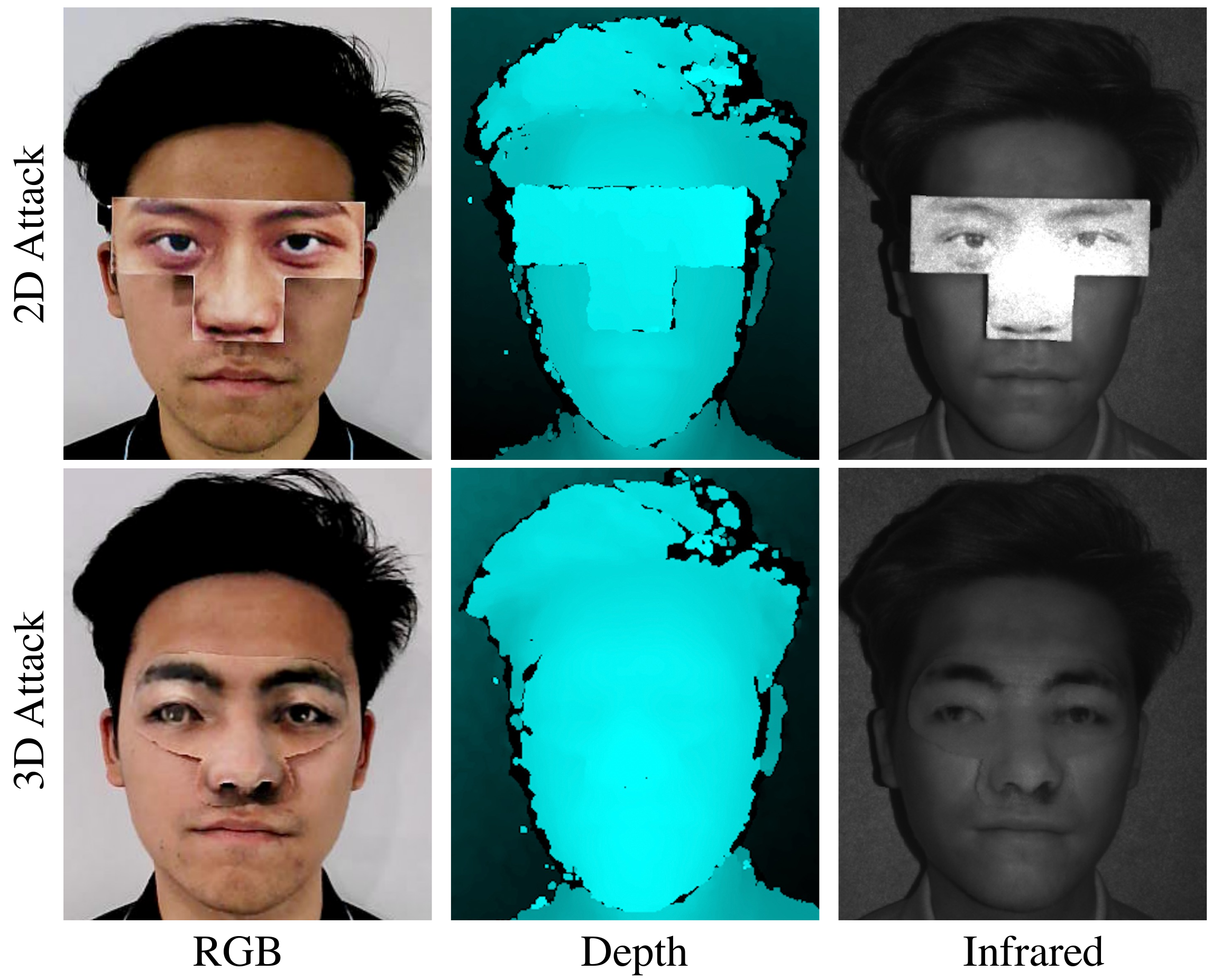}
\end{center}
\vspace{-3ex}
\caption{Visual examples of 2D and 3D attacks with three common modalities (RGB, Depth and Infrared) in face anti-spoofing.  2D attacks present intrinsic spoofing patterns among the depth and infrared modalities, which can be easily detected by the anti-spoofing detector. As a comparison, 3D attacks are more feasible for evading face anti-spoofing with multiple modalities due to versatile and realistic characteristics. }
\label{fig:spoof}
\vspace{-3ex}
\end{figure}

To evade the defensive mechanisms in the systems, we can explicitly elaborate a regional topology $\mathcal{F}'$ (as detailed in Sec.~\ref{sec:exps}) from a human face. The optimized adversarial mesh can be immediately 3D-printed and pasted on real faces for practical testing. We experimentally found that the adversarial mesh with elaborate topology can present a similar appearance with the original one among \textbf{RGB-based}, \textbf{depth-based} and \textbf{infrared-based} defensive techniques, as illustrated in Fig.~\ref{fig:spoof}. It thus becomes a more feasible way to apply the adversarial 3D mesh for physical adversarial attacks compared with 2D attacks. However, the mesh-based optimization by following the objective (\ref{eq:object}) needs to calculate gradients in high-dimensional mesh space due to thousands of points in each human face. It will easily break the geometric characteristics and surface structure of the underlying mesh manifold, thus trapping into the overfitting~\cite{liu2022imperceptible,hu2022exploring} with unsatisfactory transferability.

% the adversarial mesh by following the objective (\ref{eq:object}) is potentially incapable of achieving effective impersonation attacks against black-box recognition models, which pose an unsatisfactory black-box transferability problem. 

% the  representation space, thus inevitably resulting in a high-dimensional search space . Therefore, it is probably trapped in local optima in the optimization trajectory, leading to unsatisfactory transferability. 

% \begin{equation}
% \label{eq:rendering}
%     \bm{x}^r, \mathbf{M}=\mathrm{R}(\mathcal{M}', \bm{\rho})=\mathrm{R}(\mathcal{S},\mathcal{T},\mathcal{F}',\bm{\rho}).
% \end{equation}

% \subsection{Adversarial Textured 3D Meshes}
% Motivated by 2D adversarial patch~\cite{tong2021facesec,zhu2022masked,xiao2021improving}, we propose three practical topological structures of mesh attacks as illustrated in Fig.[].

% Eyeglass (Eye), Eyeglasses with nose (Eye\&Nose) and Respirator.

\vspace{-0.05cm}
\subsection{Low-dimensional Optimization}
\label{sec:iat3d}
\vspace{-0.05cm}
% \junz{need more elaboration/evidence on the overfitting claim.}
In this section, we aim to deviate from the existing mesh-based optimization regime, and perform the optimization trajectory in a low-dimensional manifold as a regularization for escaping from overfitting. The low-dimensional subspace must have a sufficient capacity that can encode any 3D face in this low-dimensional feature space. A principled way of spanning such a subspace is considered by leveraging 3D Morphable Model (3DMM)~\cite{blanz1999morphable}, which belongs to powerful 3D statistical models of human face shape and texture. 3DMM can effectively achieve dimensionality reduction of any high-dimensional mesh data. Therefore, optimizing in the pre-learnt low-dimensional coefficient space of 3DMM can promote more general semantic features of a 3D face. This can keep the surface structure of the underlying mesh manifold, potentially  alleviating the overfitting problem in the optimization phase.
% \junz{again, need elaboration of this overfitting issue before presenting the low-dimensional solution.}.

% To address these issues, we propose to optimize Eq.\eqref{eq:object} on a low-dimensional manifold. A low-dimensional manifold must have sufficient capacity. 

\vspace{-0.2cm}
\subsubsection{Adversarial Mesh Generation}
\vspace{-0.1cm}

Given a 2D face image $\bm{x}^{a}$, we can first predict its shape $\mathcal{S}$ and texture $\mathcal{T}$ by using 3DMM coefficients from CNN regression model\cite{deng2019accurate}, which can be represented as follows:
\begin{gather}
    \label{eq:BFM_S}
    \mathcal{S}=\overline{\mathcal{S}}+\bm{B}_{id}\bm{\alpha}+\bm{B}_{exp}\bm{\beta},\;\;\mathcal{T}=\overline{{\mathcal{T}}}+\bm{B}_{tex}\bm{\tau},
\end{gather}
where $\overline{\mathcal{S}}$ and $\overline{\mathcal{T}}$ are the averages of face shapes and textures, and $\bm{B}_{id}$, $\bm{B}_{exp}$ and $\bm{B}_{tex}$ denote the PCA bases of identity, expression and texture, respectively. Besides, a series of coefficients are regressed including $\bm{\alpha}\in\mathbb{R}^{80}$, $\bm{\beta}\in\mathbb{R}^{64}$ and
$\bm{\tau}\in\mathbb{R}^{80}$. Furthermore, this model can also regress the illumination coefficients $\bm{\gamma}\in\mathbb{R}^9$, and 
the camera position $\bm{p}\in\mathbb{R}^{6}$.
Since these coefficients are all differentiable, we thus integrate these coefficients into Eq.~\eqref{eq:object} and rewrite our objective with a variable formulation as
\begin{equation}
\small
\begin{gathered}
 \min_{\bm{\alpha}^{*}, \bm{\beta}^{*}, \bm{\tau}^{*}} \mathcal{L}_{f}(\bm{x}^{*}, \bm{x}^{b}), \text{where}\; 
    \bm{x}^{*} = \mathbf{M} \odot \bm{x}^{r} + (\bm{1} - \mathbf{M})\odot\bm{x}^{a}, \\
 \bm{x}^{r},\mathbf{M}=\mathrm{R}(\overline{\mathcal{S}}+\bm{B}_{id}\bm{\alpha}^*+\bm{B}_{exp}\bm{\beta}^*,\overline{{\mathcal{T}}}+\bm{B}_{tex}\bm{\tau}^*, \mathcal{F'}, \bm{\rho}),
\end{gathered}
\label{eq:object2}
\end{equation}
which achieves a low-dimensional optimization to make an adversarial mesh update on the parameter space of 3DMM, and we call it \textbf{AT3D-P}.

%%%%%%%%%%%%%%%%%%%%%%%%%%%%%%%%%%%%%%%%%
\begin{algorithm}[t]
\small
    \caption{Crafting Adversarial Textured Mesh}\label{algo1}
\begin{algorithmic}[1]
\Require A 3DMM model $\mathcal{G}$, a FR model ${f}$, a real face image $\bm{x}^{a}$, a target face image $\bm{x}^{b}$, the set of triangle faces $\mathcal{F}'$.
\Ensure An adversarial 3D mesh $\mathcal{M}^*$. 

\State  Get the coefficients: \{$\bm{\alpha}^a, \bm{\beta}^a, \bm{\tau}^a, {\color{blue}\bm{\gamma}^a}, {\color{blue}\bm{p}^a},\} \gets \mathcal{G}(\bm{x}^{a})$; 
\State  Get the coefficients: \{${\color{blue}\bm{\alpha}^b}, {\color{blue}\bm{\beta}^b}, {\color{blue}\bm{\tau}^b}, \bm{\gamma}^b, \bm{p}^b\} \gets \mathcal{G}(\bm{x}^{b})$;

% \State Initializing $\bm{\alpha}_0^* \gets {\color{blue}\bm{\alpha}^b}, \bm{\beta}_0^* \gets {\color{blue}\bm{\beta}^b}, \bm{\tau}_0^* \gets {\color{blue}\bm{\tau}^b}, \bm{\gamma}_0^* \gets {\color{blue}\bm{\gamma}^a}, \bm{p}_0^* \gets {\color{blue}\bm{p}^a}$;

\State Initializing $\{\bm{\alpha}_0^*, \bm{\beta}_0^*,   \bm{\tau}_0^*, \bm{\gamma}_0^*, \bm{p}_0^*\} \gets 
\{{\color{blue}\bm{\alpha}^b}, {\color{blue}\bm{\beta}^b}, {\color{blue}\bm{\tau}^b}, {\color{blue}\bm{\gamma}^a},{\color{blue}\bm{p}^a}\}$;

\For{$n$ in MaxIterations $N$}
\State \textbf{Update the coefficient $\bm{\alpha}^*$:}
\State Get \{$\mathcal{S}_{n}^{*}, \mathcal{T}_{n}^{*}$\} given \{$\bm{\alpha}_{n}^*, \bm{\beta}_n^*, \bm{\tau}_n^*$\} via Eq.~\eqref{eq:BFM_S}; 
\State Calculate $\bm{\alpha}^*_{n+1}$ via Eq.~\eqref{eq:object2} by passing \{$\mathcal{S}_{n}^{*}, \mathcal{T}_{n}^{*}$\};

\State \textbf{Update the coefficient $\bm{\beta}^*$:}
\State Get \{$\mathcal{S}_{n}^{*}, \mathcal{T}_{n}^{*}$\} given \{$\bm{\alpha}_{n+1}^*, \bm{\beta}_n^*, \bm{\tau}_n^*$\} via Eq.~\eqref{eq:BFM_S}; 
\State Calculate  $\bm{\beta}^*_{n+1}$ via Eq.~\eqref{eq:object2} by passing \{$\mathcal{S}_{n}^{*}, \mathcal{T}_{n}^{*}$\};

\State \textbf{Update the coefficient $\bm{\tau}^*$:}
\State Get \{$\mathcal{S}_{n}^{*}, \mathcal{T}_{n}^{*}$\} given \{$\bm{\alpha}_{n+1}^*, \bm{\beta}_{n+1}^*, \bm{\tau}_n^*$\} via Eq.~\eqref{eq:BFM_S}; 
\State Calculate $\bm{\tau}^*_{n+1}$ via Eq.~\eqref{eq:object2} by passing \{$\mathcal{S}_{n}^{*}, \mathcal{T}_{n}^{*}$\};

\EndFor

\State Get the shape: $\mathcal{S}^{*}\gets\overline{\mathcal{S}}+\mathbf{B}_{d}\bm{\alpha}^*_{N-1}+\mathbf{B}_{e}\bm{\beta}^*_{N-1}$
\State Get the texture: $\mathcal{T}^{*}\gets\overline{\mathcal{T}}+\mathbf{B}_{t}\bm{\tau}^*_{N-1}$
\State \Return $\mathcal{M}^{*}=(\mathcal{S}^{*}, \mathcal{T}^{*}, \mathcal{F}')$.
\end{algorithmic}
\end{algorithm}

% \textbf{Stability in the optimization.} We experimentally found that optimizing the adversarial mesh suffered from an instability problem, resulting in unsatisfactory performance. As illustrated in Fig.[], we demonstrated optimizing the adversarial outputs in the low-dimensional space can accelerate the convergence and remain stable. 

% \textbf{Overfitting problem in the optimization.} As $\epsilon$ increases, transferability indicators are also sometimes rising, but there exists a general downward trend. We aim to make adversarial mesh escape local optima for enhancing the transferability of 3D adversarial patches.

\textbf{Sensitive initialization problem.} Note that the initialization in Eq.~\eqref{eq:object2} lies in 3D mesh representation space, which is different from 2D initialization problems commonly discussed in previous works~\cite{xiao2021improving,tong2021facesec}. As presented in Table~\ref{tab:initial}, 
we found that selecting different initialization in optimizing Eq.~\eqref{eq:object2} gives rise to inconsistent black-box performances, potentially explained by falling into local optima for some cases. Thus, we apply the coefficients of the 3DMM from the victim to initialize the adversarial mesh. Note that we exploit the attacker's pose rather than the victim's one in the initialization, making the generated mesh better fit the attacker's face. 

\textbf{Optimization.}  We disturb $\bm{\alpha}^*, \bm{\beta}^*, \bm{\gamma}^*$ alternately in every attack iteration such that they can be synchronized well with each other, maintaining near their optimum during the attack. Besides, the variables can be optimized by adopting a popular optimizer, such as Adam~\cite{Kingma2014}. The detailed optimization procedure is summarized in Algorithm~\ref{algo1}.

\vspace{-0.1cm}
\subsection{Potential Advantages}
\vspace{-0.1cm}

\begin{figure*}[t]
\begin{center}
\includegraphics[width=0.9\linewidth]{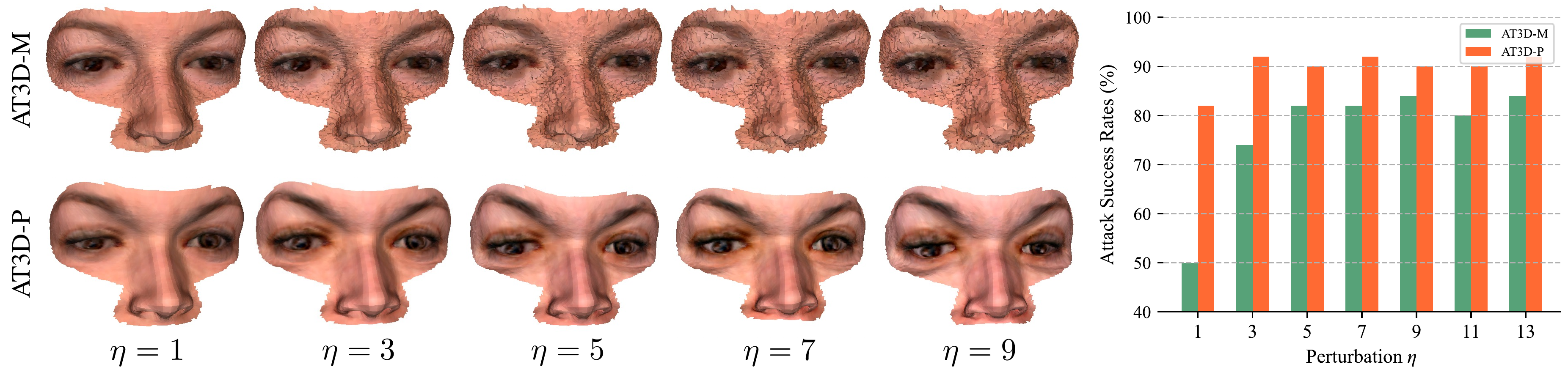}
\end{center}
\vspace{-3ex}
\caption{Experiments on how different $\eta$ affects the performance. We also further attack success rates (\%) of both attacks under different $\eta$ on LFW. MobileFace is chosen as a
white-box model, and test the performance in ResNet50. }
\vspace{-3ex}
\label{fig:epsilon}
\end{figure*}

\textbf{Naturalness.} Optimizing the coefficients of 3DMM indicates constantly searching effective linear combinations of different mesh datas, making generated adversarial mesh constrained on the data manifold of real 3D samples. As shown in Fig.~\ref{fig:epsilon}, our adversarial meshes crafted by AT3D-P appear more natural to human eyes, thus difficult to be defended by current anti-spoofing algorithms.  As a comparison, the fluctuating range of surface curves in the adversarial meshes by mesh-based optimization~\cite{xiao2019meshadv} (AT3D-M) differ significantly from those of the original samples, which also present self-intersection and flying vertices problems.

\begin{figure}[t]
\begin{center}
\includegraphics[width=0.99\linewidth]{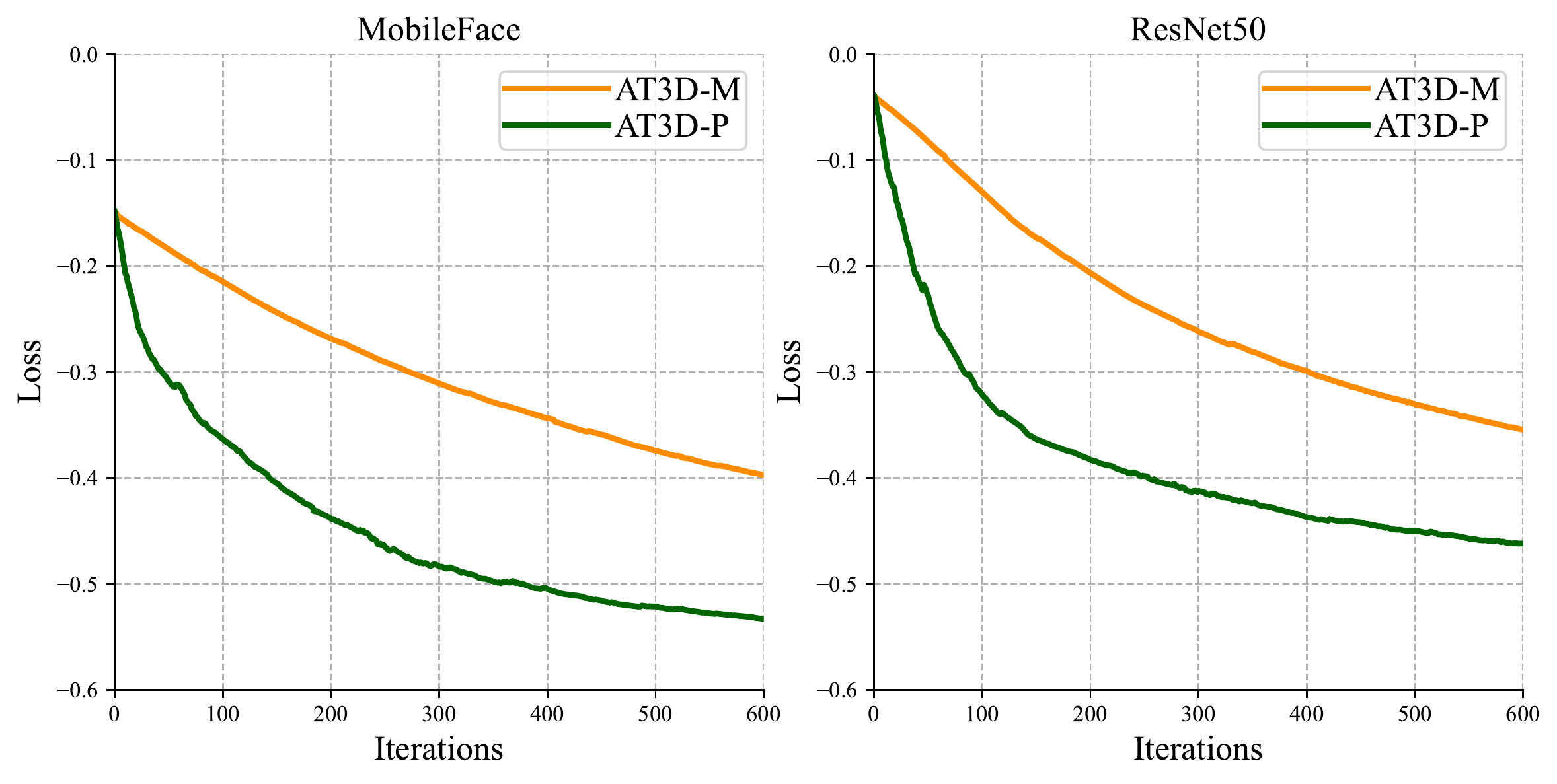}
\end{center}
\vspace{-3ex}
\caption{Comparison of loss convergence on LFW.}
\label{fig:loss}
\vspace{-3ex}
\end{figure}

\textbf{Escaping from local optima.} We experimentally found that the mesh-based optimization suffered from an inferior convergence tendency, resulting in unsatisfactory black-box performance. As illustrated in Fig.~\ref{fig:loss}, we demonstrated that optimizing the adversarial outputs in the low-dimensional space can accelerate the convergence and escape from local optima, thus achieving better transferability. Overall, AT3D-P makes a significant step toward real-world physical attack regarding naturalness and effectiveness.

%%%%%%%%%%%%%%%%%%%%%%% types %%%%%%%%%%%%
\begin{figure}[t]
\begin{center}
\includegraphics[width=0.99\linewidth]{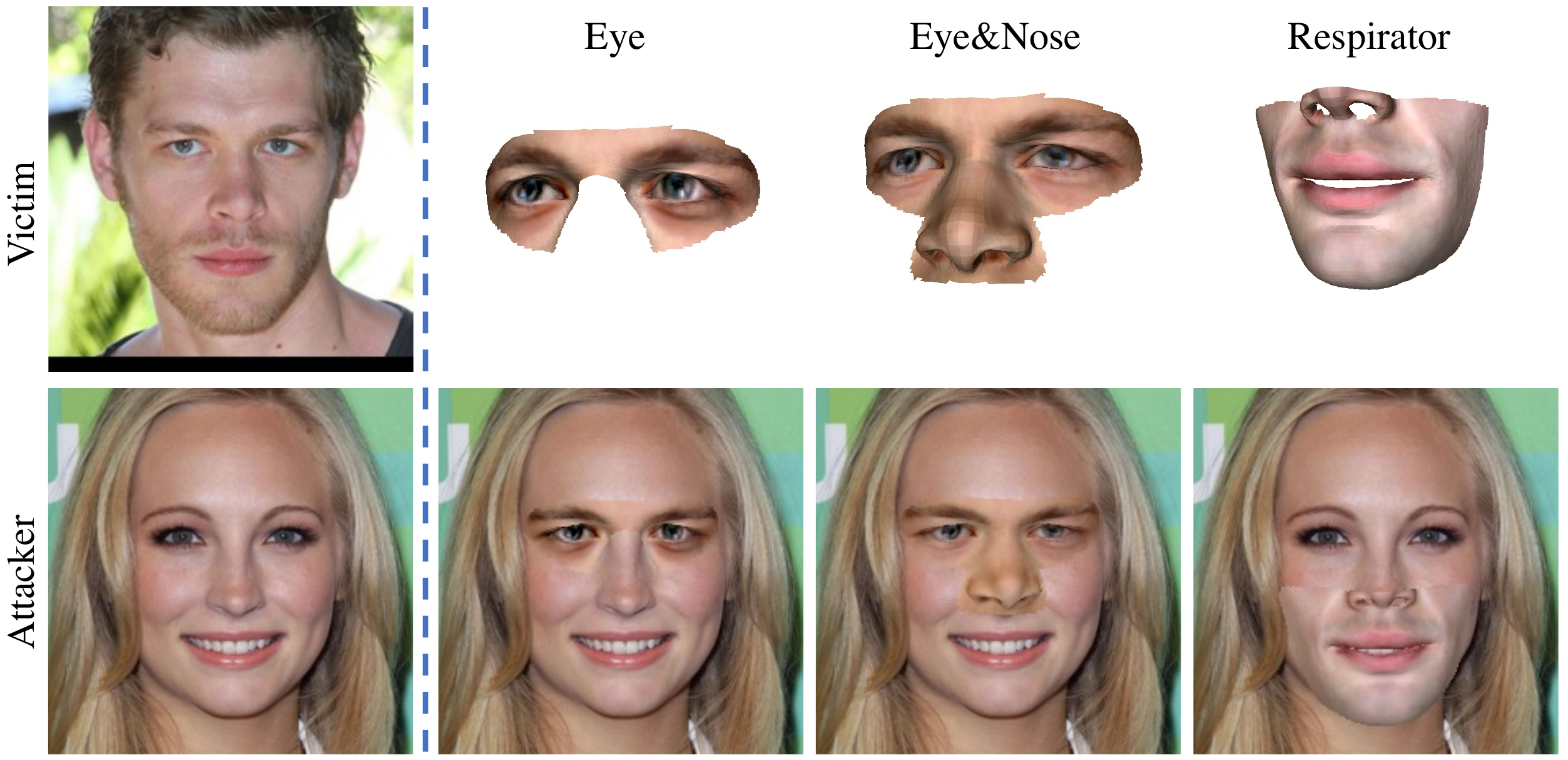}
\end{center}
\vspace{-4ex}
\caption{Three elaborate topology structures of physical adversarial attacks, including \textbf{Eye}, \textbf{Eye\&Nose} and \textbf{Respirator}.}
\label{fig:type}
\vspace{-3ex}
\end{figure}

%%%%%% visual

%%%%%%%%%%%%%%%%%
\vspace{-0.2cm}
\section{Experiments}
\vspace{-0.1cm}

\label{sec:exps}
In this section, we present the experimental results in the digital world and physical world to demonstrate the effectiveness of the proposed method.\footnote{Code at \url{https://github.com/thu-ml/AT3D}.} 
% Then we also obtain the consistent excellent performance in the physical world.

\subsection{Experiment Settings}

\textbf{Datasets.} We conduct the experiments in the digital experiments on LFW~\cite{huang2008labeled} and CelebA-HQ~\cite{karras2017progressive}, belonging to two of the most popular benchmark datasets on both low-quality and high-quality face images. For every dataset, we mainly choose 400 pairs of images with different identities to perform impersonation attacks, considering the more difficult and practical property than dodging attacks~\cite{tong2021facesec,xiao2021improving}.

\textbf{Target recognition models.} In the digital space, we study four prevailing face recognition models with different network architectures and training losses for evaluation, including ArcFace~\cite{deng2019arcface}, MobileFace~\cite{chen2018mobilefacenets}, ResNet50~\cite{he2015deep} and CosFace~\cite{wang2018cosface}. In testing, a pair of face images is fed into the model to calculate the cosine similarity (in [1, 1]), and each model can obtain over 99\% verification accuracy on LFW by following its corresponding optimal threshold. If the distance of two images exceeds the threshold, we view them as the same identities; otherwise different identities. In addition, we also evaluate the performance on three commercial face recognition APIs\footnote{Note that we do not specify which one it corresponds to in the evaluation, avoiding privacy leakage. We'll present all details in Appendix {\color{red}A}.}, \eg, Amazon, Face++, and Tencent, randomly denoted as {\color{blue}API-1}, {\color{blue}API-2}, and {\color{blue}API-3}.

\textbf{Defensive mechanisms.} We carefully studied commercial face anti-spoofing services and selected a few of the most widely used ones, such as FaceID, SenseID, Tencent and Aliyun. We randomly call them {\color{ForestGreen}D-1}, {\color{ForestGreen}D-2}, {\color{ForestGreen}D-3} and {\color{ForestGreen}D-4}.

\textbf{Physical face recognition systems.} We choose two prevailing mobile phones and two automated surveillance systems that have multiple sensors for practical testing, named {\color{blue}S-1}, {\color{blue}S-2}, {\color{blue}S-3}, and {\color{blue}S-4}. We will not disclose the manufacturer and parameters of the systems for preventing privacy leakage, only the function will be described in Appendix {\color{red}A}.

\textbf{Designed regions of mesh attack.} Motivated by 2D adversarial patch~\cite{xiao2021improving,tong2021facesec,zhu2022masked}, we propose three practical topological structures of mesh attacks as illustrated in Fig.~\ref{fig:type}, including Eyeglasses (Eye), Eyeglasses with nose (Eye\&Nose), and Respirator. We evaluate the vulnerability of face recognition models in terms of these types and compare the white-box and black-box performance.

\textbf{Compared methods.} We first choose two representative 2D methods to compare the black-box transferability, including \textbf{MIM}~\cite{Dong2017} and \textbf{EOT}~\cite{Athalye2017Synthesizing} that synthesizes examples over transformations. As for adversarial 3D meshes, we first typically craft AT3D in a mesh-based space~\cite{xiao2019meshadv}, named \textbf{AT3D-M}. Besides, multiple popular losses in mesh-based optimization, \eg, chamfer loss, laplacian loss, and edge length loss~\cite{zhang20213d}, are blended into the crafted AT3D to improve effectiveness and smoothness, named \textbf{AT3D-ML}. 

\textbf{Implementation details.} 
Note that MIM and EOT select optimal parameters as report for black-box performance  by following~\cite{xiao2021improving}. We thus set the number of iterations as $N = 400$, the learning rate $\alpha = 1.5$, the decay factor $\mu = 1$, and the size of perturbation $\epsilon = 40$ for impersonation under the $\ell_{\infty}$ norm bound. As for 3D attacks, we set the number of iterations as $N = 300$, the budget of perturbation $\eta = 3$, which belongs to a balanced choice between the effective and naturalness. These detailed hyperparameters are discussed and reported in Appendix {\color{red}A}.

%%%%%%%%%%%%%%%%% initial
\begin{table}[t]
    \begin{center}
    \small %footnotesize
    \setlength{\tabcolsep}{5pt}
    \begin{tabular}{cc|ccccc}
    \hline
      \multicolumn{2}{c|}{Initialization} &
       \multirow{2}{*}{{Res.}} & \multirow{2}{*}{{Arc.}} & \multirow{2}{*}{{Mob.}} & \multirow{2}{*}{{Cos.}} & \multirow{2}{*}{{API}} \\
       \cline{1-2}
       \textbf{Shape} & \textbf{Texture} & \\
       \hline
       \emph{Noise} & \emph{Noise} & 100.00 & 48.25 & 64.75 & 34.00 & 45.25 \\
       \emph{Attacker} & \emph{Noise} & 100.00&	43.75&	61.25&	30.50&	42.25 \\
       \emph{Attacker} & \emph{Attacker} & 100.00&	42.75&	56.50&	29.75&	41.50 \\
       \emph{Victim} & \emph{Victim} & 98.50 & 77.75 & 86.25 & 56.00 & 78.50\\
       \emph{A-Victim} & \emph{Noise} & 100.00&	79.25&	88.50&	55.50&	80.50\\
       \emph{A-Victim} & \emph{A-Victim} & 100.00&	86.25&	91.50&	61.25&	84.75 \\
       
       \hline
    \end{tabular}
    \end{center}
    \vspace{-3ex}
    \caption{Ablation study of the different initialization. ResNet50 is a white-box model.}
    \label{tab:initial}
    \vspace{-3ex}
\end{table}
%%%%%%%%%%%%%%%%

%%%%%%%%%%%%%%%% CASIA Transfer %%%%%%%%%%%%%%%%
\begin{table*}[t]
\centering
\footnotesize
\setlength{\tabcolsep}{3pt}

\begin{tabular}{c|c|ccccc|ccccc|ccccc}
\hline
\multirow{2}{*}{\begin{tabular}[c]{@{}c@{}}Source\\ Model\end{tabular}} & \multirow{2}{*}{Methods} & \multicolumn{5}{c|}{\textbf{Eye}}                                                                                                             & \multicolumn{5}{c|}{\textbf{Eye \& Nose}}                                                                                                        & \multicolumn{5}{c}{\textbf{Respirator}}                                                                                                  \\ \cline{3-17} 
&                           & \multicolumn{1}{c}{Arc.} & \multicolumn{1}{c}{Mob.} & \multicolumn{1}{c}{Cos.} & \multicolumn{1}{c}{Res.} & \multicolumn{1}{c|}{{\color{blue}API-1}} & \multicolumn{1}{c}{Arc.} & \multicolumn{1}{c}{Mob.} & \multicolumn{1}{c}{Cos.} & \multicolumn{1}{c}{Res.} & \multicolumn{1}{c|}{{\color{blue}API-1}} & \multicolumn{1}{c}{Arc.} & \multicolumn{1}{c}{Mob.} & \multicolumn{1}{c}{Cos.} & \multicolumn{1}{c}{Res.} & \multicolumn{1}{c}{{\color{blue}API-1}} \\ \hline
\multirow{4}{*}{ArcFace}   & 2D-MIM &  95.25$^*$ & 26.25 & 17.50 & 15.00 & 3.75 & \textbf{100.0}$^*$ & 66.50 & 49.25 & 49.50 & 10.75 & 91.50$^*$ & 5.50 & 8.50 & 7.50 & 2.25 \\
& 2D-EOT & \textbf{99.00}$^*$ & 49.00 & 34.50 & 35.75 & 16.25 & 99.50$^*$ & 87.75 & 73.75 & 79.00 & 36.25 & \textbf{97.75}$^*$ & 26.00 & \textbf{29.25} & 24.75 & 8.75  \\
\cline{2-17} 
& AT3D-M & 63.25$^*$ & 46.00 & 37.75 & 33.25 & 28.00 & 96.75$^*$ & 86.00 & \textbf{83.50} & 78.25 & 75.00 & 59.00$^*$ & 24.75 & 22.75 & 23.25 & 32.00\\
& AT3D-ML & 63.25$^*$ & 46.75 & 36.75 & 34.25 & 27.50 & 96.75$^*$ & 86.50 & 83.25 & 78.75 & 75.00 & 58.50$^*$ & 24.75 & 22.25 & 22.25 & 32.00 \\
& AT3D-P &	96.50$^*$&	\textbf{71.00}&	\textbf{59.00}&	\textbf{66.25}& \textbf{53.75}& \textbf{100.0}$^*$&	\textbf{95.00}&	82.00&	\textbf{93.75}&	\textbf{87.00}& 91.00$^*$&	\textbf{45.50}&	21.75&	\textbf{45.00}&	\textbf{50.25}\\ 
\hline

\multirow{4}{*}{MobileFace}   & 2D-MIM & 16.75 & 94.00$^*$ & 42.75 & 41.00 & 5.50 & 54.75 & \textbf{100.0}$^*$ & 83.25 & 82.75 & 13.00 & 18.50 & 81.50$^*$ & 22.50 & 25.75 & 1.25 \\
& 2D-EOT  &  27.75 & \textbf{100.0}$^*$ & 58.25 & 61.00 & 11.25 & 78.75 & \textbf{100.0}$^*$ & \textbf{94.50} & \textbf{96.75} & 40.00 & 32.75 & \textbf{99.50}$^*$ & \textbf{36.00} & 49.00 & 2.25 \\
\cline{2-17}
& AT3D-M & 36.00 & 71.25$^*$ & 37.75 & 35.75 & 27.25 &78.50  & 99.25$^*$ & 81.50 & 81.25 & 72.25 & 28.00 & 49.75$^*$ & 17.50 & 25.25 & 27.00 \\
& AT3D-ML & 35.25 & 71.75$^*$ & 37.50 & 35.50 & 27.25 & 79.00 & 99.25$^*$ & 81.00 & 82.00 & 73.25 & 29.00 & 50.25$^*$ & 18.25 & 25.00 & 27.00 \\
& AT3D-P & \textbf{63.75}&	98.50$^*$&	\textbf{66.75}&	\textbf{73.00}&	\textbf{52.00}&	\textbf{92.50}&	\textbf{100.0}$^*$&	87.50&	96.00&	\textbf{88.50}& \textbf{48.25}&	91.00$^*$&	21.25&	\textbf{49.75}&	\textbf{42.00}\\ 
\hline
\multirow{4}{*}{ResNet50}   & 2D-MIM & 13.75 & 40.50 & 35.50 & 93.25$^*$ & 3.50 & 53.25 & 88.25 & 76.25 & \textbf{100.0}$^*$ & 13.25 & 18.50 & 21.50 & 23.00 & 85.00$^*$ & 1.75 \\
& 2D-EOT  & 20.50 & 65.00 & 48.50 & \textbf{100.0}$^*$ & 13.25 & 72.50 & \textbf{96.25} & \textbf{86.50} & \textbf{100.0}$^*$ & 43.00 & 34.50 & 49.75 & \textbf{36.25} & \textbf{99.00}$^*$ & 4.25 \\
\cline{2-17} 
& AT3D-M & 32.75 & 44.75 & 35.25 & 65.00$^*$ & 26.50 &  74.75 & 85.00 & 76.75 & 97.00$^*$ & 71.25 & 28.50 & 23.00 & 17.25 & 48.75$^*$ & 26.50 \\
& AT3D-ML & 34.00 & 44.50 & 34.75 & 65.25$^*$ & 27.25 &  74.50 & 84.50 & 75.50 & 97.00$^*$ & 70.50 & 28.00 & 23.50 & 18.00 & 47.75$^*$ & 26.50 \\
& AT3D-P &\textbf{59.75}&	\textbf{74.75}&	\textbf{56.25}&	99.00$^*$&	\textbf{52.50}&\textbf{92.00}&	\textbf{96.25}&	78.75&	\textbf{100.0}$^*$&	\textbf{88.50} & \textbf{46.00}&	\textbf{52.00}&	20.75&	91.25$^*$&	\textbf{44.25}\\ 
\hline

\end{tabular}
\vspace{-2ex}
\caption{The attack success rates ($\%$) of  the face recognition models on CelebA-HQ with adversarial meshes. $^*$ indicates white-box attacks.
}
\label{tab:transfer-CelebAHQ}
\vspace{-1ex}
\end{table*}

%%%%%%%%%%%%%%%% CASIA Ablation %%%%%%%%%%%%%%%%

\begin{table*}[t]
\centering
\footnotesize
\setlength{\tabcolsep}{3pt}

\begin{tabular}{c|c|ccccc|ccccc|ccccc}
\hline
\multirow{2}{*}{\begin{tabular}[c]{@{}c@{}}Source\\ Model\end{tabular}} & \multirow{2}{*}{Methods} & \multicolumn{5}{c|}{\textbf{Eye}}                                                                                                             & \multicolumn{5}{c|}{\textbf{Eye \& Nose}}                                                                                                        & \multicolumn{5}{c}{\textbf{Respirator}}                                                                                                  \\ \cline{3-17} 
&           & \multicolumn{1}{c}{Arc.} & \multicolumn{1}{c}{Mob.} & \multicolumn{1}{c}{Cos.} & \multicolumn{1}{c}{Res.} & \multicolumn{1}{c|}{{\color{blue}API-1}} & \multicolumn{1}{c}{Arc.} & \multicolumn{1}{c}{Mob.} & \multicolumn{1}{c}{Cos.} & \multicolumn{1}{c}{Res.} & \multicolumn{1}{c|}{{\color{blue}API-1}} & \multicolumn{1}{c}{Arc.} & \multicolumn{1}{c}{Mob.} & \multicolumn{1}{c}{Cos.} & \multicolumn{1}{c}{Res.} & \multicolumn{1}{c}{{\color{blue}API-1}} \\ \hline
\multirow{3}{*}{ArcFace} 
& \{$\bm{\alpha}$, $\bm{\beta}$\}  & 77.75$^*$&	57.75&	53.00&	47.75&	47.75&	98.25$^*$&	89.75&	77.50&	82.50&	84.25& 67.25$^*$&	32.00&	17.75&	31.00&	39.50\\
& \{$\bm{\tau}$\} & 86.50$^*$&	57.00&	53.25&	51.00 & 41.50&	98.50$^*$&	89.00&	73.50&	98.50 & 77.50& 73.25$^*$&	34.00&	19.75&	33.50&	42.50\\
& \{$\bm{\alpha}$, $\bm{\beta}$, $\bm{\tau}$\} &	\textbf{96.50}$^*$&	\textbf{71.00}&	\textbf{59.00}&	\textbf{66.25}& \textbf{53.75}& \textbf{100.0}$^*$&	\textbf{95.00}&	\textbf{82.00}&	\textbf{93.75}&	\textbf{87.00}& \textbf{91.00}$^*$&	\textbf{45.50}&	\textbf{21.75}&	\textbf{45.00}&	\textbf{50.25}\\ 
\hline

\multirow{3}{*}{MobileFace}
& \{$\bm{\alpha}$, $\bm{\beta}$\} & 49.25&	83.25$^*$&	52.75&    52.50& 43.00&	83.25&	99.00$^*$&	77.75&	88.00  &	81.75& 34.00&	69.25$^*$&	17.50&	29.50&	32.00\\
& \{$\bm{\tau}$\} & 50.75&	90.00$^*$&	57.75&		59.00 & 43.00&	85.75&	99.75$^*$&	77.00&	89.50 &	78.00& 38.25&	70.00$^*$&	20.25&	34.50&	37.50\\
& \{$\bm{\alpha}$, $\bm{\beta}$, $\bm{\tau}$\} & \textbf{63.75}&	\textbf{98.50}$^*$&	\textbf{66.75}&	\textbf{73.00}&	\textbf{52.00}&	\textbf{92.50}&	\textbf{100.0}$^*$&	\textbf{87.50}&	\textbf{96.00}&	\textbf{88.50}& \textbf{48.25}&	\textbf{91.00}$^*$&	\textbf{21.25}&	\textbf{49.75}&	\textbf{42.00}\\ 
\hline
\multirow{3}{*}{ResNet50} & \{$\bm{\alpha}$, $\bm{\beta}$\} & 43.25&	55.75&	47.25&	86.00$^*$&	43.50&	82.00&	89.25&	75.25&	98.50$^*$&	80.25 & 32.25&	32.75&	18.00&	67.50$^*$&	34.00\\
& \{$\bm{\tau}$\} & 44.50&	59.50&	49.75&	89.50$^*$&	42.50&	79.75&	88.50&	72.00&	99.00$^*$&	78.00 & 34.50&	34.00&	18.50&	73.00$^*$&	36.00\\
& \{$\bm{\alpha}$, $\bm{\beta}$, $\bm{\tau}$\} &\textbf{59.75}&	\textbf{74.75}&	\textbf{56.25}&	\textbf{99.00}$^*$&	\textbf{52.50}&\textbf{92.00}&	\textbf{96.25}&	\textbf{78.75}&	\textbf{100.0}$^*$&	\textbf{88.50} & \textbf{46.00}&	\textbf{52.00}&	\textbf{20.75}&	\textbf{91.25}$^*$&	\textbf{44.25}\\ 

\hline
\end{tabular}
\vspace{-2ex}
\caption{The attack success rates ($\%$) of different coefficients on CelebA-HQ with adversarial meshes. $^*$ indicates white-box attacks.}
\label{tab:ablation-CelebAHQ}
\vspace{-3ex}
\end{table*}

%%%%%%%%%%%%%%%%%%%
\subsection{Black-box Attacks in the Digital World}
In this section,  we present the experimental results of white-box and black-box attacks in the digital world. Specifically, we consider three practical topological structures of mesh attacks. Due to the space limitation, we only present the evaluation results on CelebA-HQ in this section, and the results on LFW are listed in Appendix {\color{red}B}.
% , which present similar overall conclusion.

\begin{figure*}[t]
\begin{center}
\includegraphics[width=0.95\linewidth]{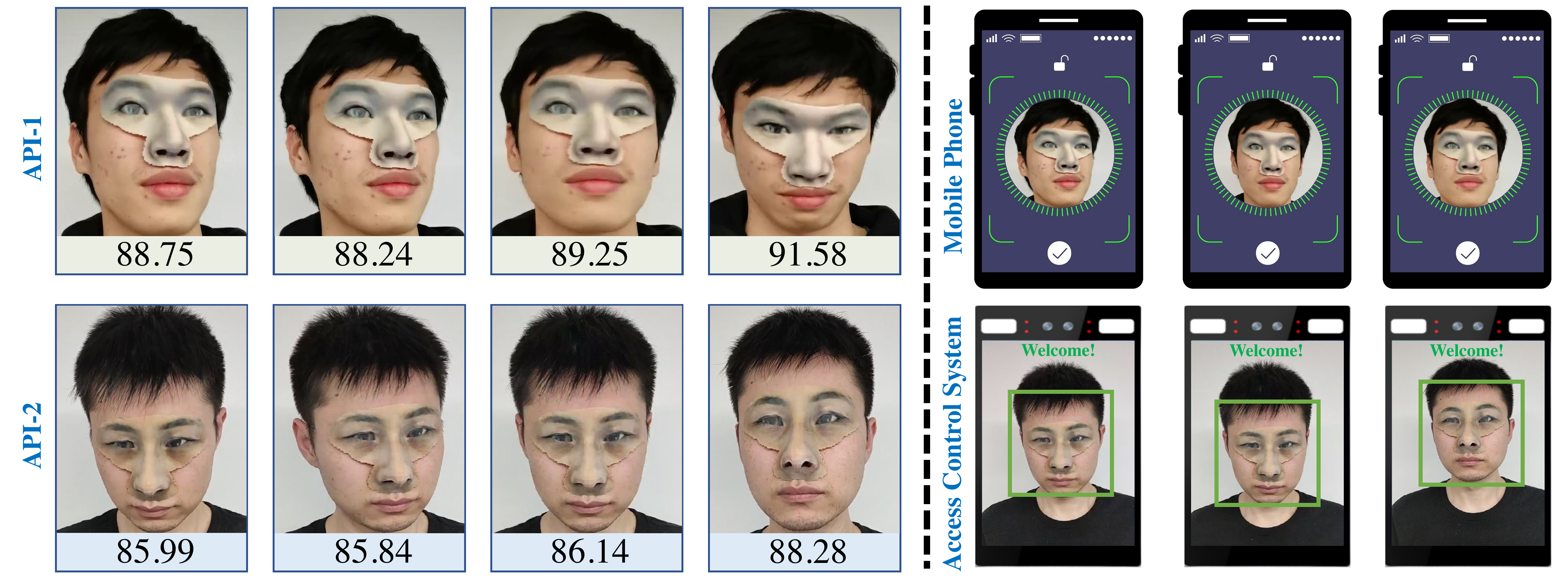}
\end{center}
\vspace{-4ex}
\caption{Experimental results of physical attacks by wearing the 3D adversarial meshes, which can achieve effective impersonation attacks on two recognition APIs, one mobile phone and one automated access control system. }
%Note that the two APIs obtain very low similarities of less than 20\% when identifying the attackers as the corresponding target identities at first.
% After wearing the adversarial meshes, the attackers can successfully impersonate the target identities.
\label{fig:phy}
\vspace{-2ex}
\end{figure*}
%%%%%%%%%%%%%%%%%%%%%%%

\textbf{Effectiveness of the proposed method.}
To verify the effects of different settings, we compare the performance of different methods. Table~\ref{tab:transfer-CelebAHQ} show the attack success rates ($\%$) of the different face recognition models. Although 2D attacks obtain effective white-box performance, yet failing to steadily present acceptable black-box transferability. Besides, 2D attacks present intrinsic spoofing patterns in Fig.~\ref{fig:spoof}, which are easily detectable by face anti-spoofing. As for 3D attacks, AT3D-ML can enhance smoothness by using multiple losses in mesh-based optimization (as visually presented in Appendix {\color{red}C}). However, we found that these losses cannot promote more transferable adversarial meshes. As a whole, AT3D-P can obviously obtain the best black-box attack success of face recognition models among all 2D and 3D attacks in most testing settings. The reason is that AT3D-P fully leverages low-dimensional optimization based on 3DMM, making generated adversarial meshes more effective and transferable for black-box models. In addition, we will have priority to select Eye\&Nose for conducting practical attacks considering its effectiveness.

\textbf{Better visual quality.} To further examine the naturalness of crafted adversarial samples, we perform experiments with different $\eta$. Fig.~\ref{fig:epsilon} shows the evaluation results of AT3D-M and AT3D-P  w.r.t naturalness and black-box transferability. As $\eta$ increases, the generated meshes of AT3D-M present worse visual quality, and expose flying vertices problems. As a comparison, AT3D-P can consistently obtain smooth appearances meanwhile acquiring better attack success rates. This tendency is also verified by common distances of evaluating naturalness, as detailed in Appendix~{\color{red}C}.

\vspace{-0.3cm}
\subsubsection{Ablation Study}
\vspace{-0.1cm}
\textbf{Initialization.} In Table~\ref{tab:initial}, we exploit different initialization to demonstrate the effectiveness in the shape and texture space, e,g., pasting uniform noises, original attacker and victim image. Adopting the initialization of the victim performs better among all black-box testings. Furthermore, the best performance can be achieved when adopting the pose of the victim to consistently fit the attacker's face, denoted as A-Victim. The semantic feature between the victim’s face and the final crafted mesh is usually closer than that between the random noise and the final mesh. Thus an adversary would prefer to accelerate the optimization process and potentially alleviate overfitting by benefiting from the initialization of the victim. 

% We thus adopt this initialization to craft adversarial meshes as default.

% We can see that typical initialization fails to exhibit acceptable transferability potentially due to overfitting in local optima, whereas 

\textbf{Coefficients of 3DMM.} We conduct an ablation study as shown in Table~\ref{tab:ablation-CelebAHQ} to investigate the coefficients of 3DMM. Optimizing the coefficients $\{\bm{\alpha}, \bm{\beta}, \bm{\tau}\}$ can obtain a better effective performance than its subsets $\{\bm{\alpha}, \bm{\beta}\}$ and $\{\bm{\tau}\}$. This also indicates that adversarial meshes benefit from texture and shape space in the optimization phase, making them more effective in white-box and black-box testing.

\subsection{Experiments in the Physical World}

In this section, we conduct \textbf{50} attacker-to-victim pairs to conduct the experiments to verify the effectiveness of the proposed method in the physical world. The procedure is evaluated by: 1) taking a face photo of a volunteer with a fixed camera under natural light; 2) crafting adversarial textured meshes for each volunteer; 3) achieving 3D printing and pasting them on real faces of the volunteers; 4) testing the attack performance against practical face recognition system. We also provide 3D-printed techniques and testing details in Appendix {\color{red}D}.

\textbf{Misleading commercial recognition APIs.} The working mechanism and training data in APIs are completely known for us. As illustrated in Table~\ref{tab:api} and Fig.~\ref{fig:phy}, black-box APIs obtain very low similarity when identifying the attackers as the corresponding target identities at first. After wearing the adversarial meshes, the attackers can successfully impersonate the target identities, as predicted by the model. These results illustrate the effectiveness of our method against the commercial face recognition APIs. The main reason is that AT3D benefits from appropriate topology and effective optimization, and presents consistent effectiveness in the both digital and real world.

\textbf{Bypassing defensive mechanisms.} To verify the effectiveness of AT3D-P in face anti-spoofing, we choose several strong commercial face anti-spoofing APIs. The crafted adversarial images by AT3D-P will be fed into the black-box API for evaluating the performance. As shown in Table~\ref{tab:api}, we can obtain a steady performance on passing the face anti-spoofing API with a high success rate. Thus 3D attacks are also conducive to passing commercial face anti-spoofing due to realistic and versatile characteristics.

\textbf{Evaluation on practical commercial systems.}
We further conduct physical experiments on multiple commercial systems, including prevailing mobile phones and automated surveillance systems. For the device {\color{blue}S-1}, we can easily import the victims' information in batches into the system when achieving attacker-to-victim adversarial testing. For {\color{blue}S-2}, {\color{blue}S-3} and {\color{blue}S-4}, we only import every victim's information in sequence into the system. Considering the limited resources and complicated procedures for these three devices, we randomly choose \textbf{10} pairs to conduct these experiments. As shown in Table~\ref{tab:sys}, our method also obtains consistent effective performance in these challenging devices. We will present all detailed results in Appendix {\color{red}D} for every testing pair against different recognition systems.
%%%%%%physical
\begin{table}[t]
    \begin{center}
    \small %footnotesize
    \setlength{\tabcolsep}{3pt}
    \begin{tabular}{c|ccc|cccc}
    \hline
      & \multicolumn{3}{c|}{\textbf{Face Recognition}}
       & \multicolumn{4}{c}{\textbf{Face Anti-spoofing}} \\
       \cline{2-8}
        & {\color{blue}API-1} & {\color{blue}API-2} & {\color{blue}API-3} & {\color{ForestGreen}D-1} & {\color{ForestGreen}D-2} & {\color{ForestGreen}D-3} & {\color{ForestGreen}D-4}\\
        \hline
       Origin & 22.21 & 8.50 & 24.09 & -& - & - & -\\
       AT3D & 82.45 & 84.20& 74.87 & \textbf{46}/50 & \textbf{48}/50 & \textbf{41}/50 & \textbf{48}/50\\ 
       $\bm{\Delta}$ & {\color{myred}\textbf{$+$60.24}} & {\color{myred}\textbf{$+$75.70}} & {\color{myred}\textbf{$+$50.78}}& -& - & - & -\\
       \hline
    \end{tabular}
    \end{center}
    \vspace{-4ex}
    \caption{The mean similarity (\%) or passing number of \textbf{50} physical pairs with printed adversarial meshes against APIs.}
    \label{tab:api}
    \vspace{-1ex}

\end{table}
%%%%%%%%%%%%%%%%

%%%%%%physical2
\begin{table}[t]
    \begin{center}
    % \small %footnotesize
    \setlength{\tabcolsep}{7pt}
    \begin{tabular}{c|cccc}
    \hline
      
        & {{\color{blue}S-1}} & {{\color{blue}S-2}} & {{\color{blue}S-3}} &{{\color{blue}S-4}}\\
        
        \hline
      Physical Evaluation & \textbf{23}/50 & \textbf{6}/10 & \textbf{7}/10 & \textbf{3}/10 \\
       \hline
    \end{tabular}
    \end{center}
    \vspace{-3ex}
    \caption{The passing number of printed adversarial meshes against the practical systems that achieved face recognition and defense.}
    \label{tab:sys}
    \vspace{-3ex}

\end{table}
%%%%%%%%%%%%%%%%
% \vspace{-0.1cm}
\section{Conclusion}
% \vspace{-0.1cm}
In this paper, we developed effective and practical adversarial textured 3D meshes with an elaborate topology to evade the defenses. Besides, we proposed to perturb the low-dimensional coefficients from 3DMM, which significantly improves black-box transferability meanwhile obtaining faster search efficiency and better visual quality.
Extensive experiments demonstrate that our method can consistently mislead multiple commercial recognition systems.

\vspace{-0.1cm}
\section*{Acknowledgements}

\footnotesize{
This work was supported by the National Key Research and Development Program of China 2020AAA0106302, NSFC Projects (Nos. 62276149, 61620106010, 62076147, U19A2081, U19B2034, U1811461), 
Alibaba Group through Alibaba Innovative Research Program, Tsinghua-Huawei Joint Research Program, a
grant from Tsinghua Institute for Guo Qiang, Tsinghua-OPPO Joint Research Center for Future Terminal Technology, and the High Performance Computing Center, Tsinghua University. Y. Dong was also supported by the China National Postdoctoral Program for Innovative Talents and Shuimu Tsinghua Scholar Program. C. Liu and L. Xu were intern at RealAI during this work. J. Zhu was also supported by the XPlorer Prize.}

%%%%%%%%% REFERENCES
{\small
\bibliographystyle{ieee_fullname}
\bibliography{egbib}
}

% \clearpage

\appendix

\section{Implementation Details}

\subsection{The Details of Commercial Services}
As illustrated in Table~\ref{tableappendix1}, we provide the public links of APIs and the public information of practical systems used in our paper. 

\begin{itemize}
    \item[1)] \textbf{Face recognition APIs}:
    We consider three popular face recognition APIs, including Amazon, Tencent, Face++. They are widely used in financial payment and automated surveillance systems.

    \item[2)] \textbf{Face anti-spoofing APIs}: Currently, the mainstream face anti-spoofing solutions include cooperative living detection and non-cooperative living detection (silent living detection). Cooperative living detection requires the user to complete some specified action according to the prompt, and then output the live verification. As a comparison, silent live detection directly aims to directly judge whether the face in front of the machine is real or fake, which is widely used and studied due to convenience and practicability for users. Therefore, we mainly focus on silent living detection in this paper, \eg, FaceID, SenseID, considering their practicability and \emph{millions of} API calls every day.

    \item[3)] \textbf{Practical systems}: We choose two prevailing mobile phones and two automated surveillance systems that have multiple sensors for practical testing as described in Table~\ref{tableappendix1}. Note that we did not disclose the manufacturer and parameters of the practical systems for preventing privacy leakage.

\end{itemize}

\subsection{Some Hyperparameters}

For 3D attacks, we set the number of iterations as $N = 300$, the budget of perturbation $\eta = 3$, which belongs to a balanced choice between the effective and naturalness. Besides, we utilize
Adam optimizer and set the initial learning rate as $1.5 * \eta / N$.

% Please note that the checkpoints of evaluating adversarial robustness are released in their code implementations.

\begin{table*}[htbp]
  \centering
  \small
  \vspace{-0.cm}
  \setlength{\tabcolsep}{5pt}

    \begin{tabular}{c|c|l}
    \hline
   & Name & Public link / information \\
   \hline
  \multirow{3}{*}{Face Recognition} & Amazon & \url{https://aws.amazon.com/rekognition/}\\
  & Face++ & \url{https://www.faceplusplus.com.cn/face-comparing/}\\
  & Tencent & \url{https://www.tencentcloud.com/products/facerecognition}\\
  \hline
  \multirow{4}{*}{Anti-Spoofing} & FaceID & \url{https://www.faceplusplus.com.cn/faceid-solution/}\\
  & SenseID & \url{https://www.sensetime.com/senseid/}\\
  & Tencent & \url{https://www.tencentcloud.com/products/facerecognition}\\
  & Aliyun & \url{https://s.alibaba.com/product/cloudauth/}\\ 
      
    \hline
\multirow{4}{*}{Practical System} & Mobile Phone 1 & including RGB, depth cameras, supporting recognition and anti-spoofing \\
& Mobile Phone 2 & including RGB, depth cameras, supporting recognition and anti-spoofing\\
& Access Control System 1 & including RGB, depth and infrared cameras, supporting recognition and anti-spoofing\\
& Access Control System 2 & including RGB, depth and infrared cameras, supporting recognition and anti-spoofing\\
    
    \hline
    \end{tabular}%
    \caption{We list the public links of APIs or public information of practical systems used in this paper.}
    \label{tableappendix1}
\end{table*}%

\section{More Experiments}

Table~\ref{tab:transfer-LFW} show the attack success rates ($\%$) of the different face recognition models on the LFW dataset. We also conduct an ablation study as shown in Table~\ref{tab:ablation-LFW} on LFW to fully investigate the coefficients of 3DMM.
As illustrated in the evaluation results on CelebA-HQ, the effects on LFW  present a similar overall conclusion. As a whole, AT3D-P can obviously obtain the best black-box attack success of face recognition models among all 2D and 3D attacks in most testing settings. The reason is that AD3D-P fully leverages low-dimensional optimization based on 3DMM, making generated adversarial meshes more effective and transferable for black-box models.
% We show the four mentioned types of physically realizable adversarial attacks in this paper in Fig.[]

%%%%%%%%%%%%%%%% LFW Transfer %%%%%%%%%%%%%%%%
\begin{table*}[t]
\centering
\footnotesize
\setlength{\tabcolsep}{3pt}

\begin{tabular}{c|c|ccccc|ccccc|ccccc}
\hline
\multirow{2}{*}{\begin{tabular}[c]{@{}c@{}}Source\\ Model\end{tabular}} & \multirow{2}{*}{Methods} & \multicolumn{5}{c|}{\textbf{Eye}}                                                                                                             & \multicolumn{5}{c|}{\textbf{Eye \& Nose}}                                                                                                        & \multicolumn{5}{c}{\textbf{Respirator}}                                                                                                  \\ 
\cline{3-17} 
&                           & \multicolumn{1}{c}{Arc.} & \multicolumn{1}{c}{Mob.} & \multicolumn{1}{c}{Cos.} & \multicolumn{1}{c}{Res.} & \multicolumn{1}{c|}{{\color{blue}API-1}} & \multicolumn{1}{c}{Arc.} & \multicolumn{1}{c}{Mob.} & \multicolumn{1}{c}{Cos.} & \multicolumn{1}{c}{Res.} & \multicolumn{1}{c|}{{\color{blue}API-1}} & \multicolumn{1}{c}{Arc.} & \multicolumn{1}{c}{Mob.} & \multicolumn{1}{c}{Cos.} & \multicolumn{1}{c}{Res.} & \multicolumn{1}{c}{{\color{blue}API-1}} \\ \hline
\multirow{4}{*}{ArcFace}   & 2D-MIM  &  90.50$^*$ & 15.50 & 13.50 & 8.50 & 2.00 & 99.25$^*$ & 52.50 & 35.25 & 34.50 & 4.50 & 87.00$^*$ & 6.75 & 7.00 & 5.00 & 1.50 \\
& 2D-EOT  &  \textbf{98.75}$^*$ & 37.50 & 30.00 & 22.75 & 10.00 & 99.50$^*$ & 80.25 & 60.50 & 65.75 & 19.25 & \textbf{98.25}$^*$ & 21.75 & \textbf{22.50} & 17.50 & 5.50\\
\cline{2-17} 
& AT3D-M & 46.00$^*$ & 29.25 & 27.75 & 21.75 & 22.50 & 89.75$^*$ & 68.00 & 67.50 & 60.25 & 64.25 & 44.50$^*$ & 18.25 & 19.50 & 13.75 & 29.25 \\
& AT3D-ML & 46.00$^*$ & 29.00 & 28.50 & 21.00 & 22.25 & 91.00$^*$ & 67.75 & \textbf{68.50} & 60.75 & 65.00 & 45.50$^*$ & 18.75 & 20.00 & 14.00 & 28.50 \\
& AT3D-P &93.75$^*$& \textbf{59.00}&	\textbf{41.00}&	\textbf{52.25}&	\textbf{47.50} & \textbf{99.75}$^*$&	\textbf{89.75}&	{59.25}& \textbf{86.50}& \textbf{85.25} & 89.00$^*$&	\textbf{41.00}&	{11.50}&	\textbf{39.25}&	\textbf{45.25}\\ 
\hline
\multirow{4}{*}{MobileFace}   & 2D-MIM  & 6.50 & 91.25$^*$ & 30.50 & 28.75 & 3.25 & 35.75 & \textbf{100.0}$^*$ & 62.75 & 70.25 & 7.50 & 12.75 & 80.75$^*$ & 19.00 & 19.00 & 1.50\\
& 2D-EOT  & 14.25 & \textbf{100.0}$^*$ & \textbf{48.25} & 50.50 & 6.25 & 59.50 & \textbf{100.0}$^*$ & \textbf{82.25} & \textbf{90.25} & 18.50 & 19.75 & \textbf{99.50}$^*$ & \textbf{38.00} & 45.00 & 2.00  \\
\cline{2-17} 
& AT3D-M & 21.50 & 58.50$^*$ & 28.25 & 25.75 & 21.75 & 62.50 & 93.75$^*$ & 61.75 & 64.50 & 63.25 & 21.50 & 45.25$^*$ &  19.25 & 14.75 & 26.75 \\
& AT3D-ML & 21.50 & 58.00$^*$ & 27.50 & 26.25 & 21.50 & 63.75 & 93.25$^*$ & 61.75 & 65.50 & 63.25 & 21.75 & 45.75$^*$ & {19.25} & 15.25 & 26.75 \\

& AT3D-P & \textbf{50.50}&	85.75$^*$&	{42.50}&	\textbf{54.25}&	\textbf{41.00} & \textbf{82.25}&	95.75$^*$&	{63.00}&	{82.75}&	\textbf{82.00} & \textbf{39.50}&	91.00$^*$& {11.25}&	\textbf{46.00}&	\textbf{39.25}\\
\hline
\multirow{4}{*}{ResNet50}   & 2D-MIM  & 5.50 & 33.50 & 27.50 & 89.25$^*$ & 3.00 & 41.00 & 82.50 & 64.50 & 99.75$^*$ & 7.00 & 13.75 & 27.50 & 19.75 & 85.25$^*$ & 2.00 \\
& 2D-EOT  & 13.00 & 55.25 & 39.00 & \textbf{99.50}$^*$ & 9.25 & 63.75 & \textbf{94.75} & \textbf{79.75} & \textbf{100.0}$^*$ & 33.50 & 19.75 & \textbf{48.00} & \textbf{34.00} & \textbf{99.00}$^*$ & 5.25 \\
\cline{2-17} 
& AT3D-M & 19.50 & 30.50 & 24.25 & 47.25$^*$ &  20.75 & 58.50 & 68.75 & 59.50 & 91.75$^*$ & 62.75 & 18.00 & 19.50 & 17.25 & 35.50$^*$ & 25.00\\
& AT3D-ML & 19.75 & 29.75 & 24.75 & 47.50$^*$ & 20.50 & 57.25 & 68.00 & 59.75 & 91.25$^*$ & 62.25 & 18.25 & 18.50 & 16.75 & 35.50$^*$ & 24.50 \\
& AT3D-P &\textbf{48.00}& \textbf{62.75}&	\textbf{42.50}&	95.00$^*$&	\textbf{47.50}&\textbf{86.25}&	{91.50}&	{61.25}&	\textbf{100.00}$^*$& \textbf{84.75}& \textbf{33.75}&	{44.75}&	{11.75}&	90.50$^*$& \textbf{43.25}\\
\hline
\end{tabular}
% \vspace{-2ex}
\caption{The attack success rates ($\%$) of  the face recognition models on LFW with adversarial meshes. $^*$ indicates white-box attacks.}
\label{tab:transfer-LFW}
% \vspace{-2ex}
\end{table*}

%%%%%%%%%%%%%%%% LFW Ablation %%%%%%%%%%%%%%%%
\begin{table*}[t]
\centering
\footnotesize
\setlength{\tabcolsep}{3pt}
\begin{tabular}{c|c|ccccc|ccccc|ccccc}
\hline
\multirow{2}{*}{\begin{tabular}[c]{@{}c@{}}Source\\ Model\end{tabular}} & \multirow{2}{*}{Settings} & \multicolumn{5}{c|}{\textbf{Eye}} & \multicolumn{5}{c|}{\textbf{Eye \& Nose}} & \multicolumn{5}{c}{\textbf{Respirator}}  \\ 
\cline{3-17} 
&                           & \multicolumn{1}{c}{Arc.} & \multicolumn{1}{c}{Mob.} & \multicolumn{1}{c}{Cos.} & \multicolumn{1}{c}{Res.} & \multicolumn{1}{c|}{{\color{blue}API-1}} & \multicolumn{1}{c}{Arc.} & \multicolumn{1}{c}{Mob.} & \multicolumn{1}{c}{Cos.} & \multicolumn{1}{c}{Res.} & \multicolumn{1}{c|}{{\color{blue}API-1}} & \multicolumn{1}{c}{Arc.} & \multicolumn{1}{c}{Mob.} & \multicolumn{1}{c}{Cos.} & \multicolumn{1}{c}{Res.} & \multicolumn{1}{c}{{\color{blue}API-1}} \\ \hline
\multirow{3}{*}{ArcFace} 
& \{$\bm{\alpha}$, $\bm{\beta}$\}  & 66.50$^*$&	43.50&	35.50&	34.75&	35.75 & 94.00$^*$ & 77.25 & 57.00 &	72.00 & 75.50 & 62.00$^*$&	25.75&	9.25&	21.75&	32.50\\
& \{$\bm{\tau}$\} & 72.50$^*$&	43.50&	38.50&	39.00&	35.75 & 95.50$^*$ & 78.00 &	57.00  & 72.50 & 72.75 & 61.25$^*$&	25.50&	11.75&	21.50&	34.75\\
& \{$\bm{\alpha}$, $\bm{\beta}$, $\bm{\tau}$\} &	\textbf{93.75}$^*$&	\textbf{59.00}&	\textbf{41.00}&	\textbf{52.25}&	\textbf{47.50} & \textbf{99.75}$^*$&	\textbf{89.75}&	\textbf{59.25}& \textbf{86.50}& \textbf{85.25} & \textbf{89.00}$^*$&	\textbf{41.00}&	\textbf{11.50}&	\textbf{39.25}&	\textbf{45.25}\\ 
\hline

\multirow{3}{*}{MobileFace}
& \{$\bm{\alpha}$, $\bm{\beta}$\}  &36.75&	74.75$^*$&	39.75&	39.75&	36.00 & 76.00&	97.00$^*$&	58.50&	74.50&	78.00 & 24.75&	61.75$^*$&	8.00&	26.75&	26.50\\
& \{$\bm{\tau}$\} & 39.50&	83.50$^*$&	39.75&	44.75&	38.00 & 72.25&	98.50$^*$&	61.75&	77.50&	76.50 & 27.25&	63.50$^*$&	10.50&	30.50&  35.75\\
& \{$\bm{\alpha}$, $\bm{\beta}$, $\bm{\tau}$\} & \textbf{50.50}&	\textbf{85.75}$^*$&	\textbf{42.50}&	\textbf{54.25}&	\textbf{41.00} & \textbf{82.25}&	\textbf{95.75}$^*$&	\textbf{63.00}&	\textbf{82.75}$^*$&	\textbf{82.00} & \textbf{39.50}&	\textbf{91.00}$^*$&	\textbf{11.25}&	\textbf{46.00}&	\textbf{39.25}\\ 
\hline
\multirow{3}{*}{ResNet50} & \{$\bm{\alpha}$, $\bm{\beta}$\} & 31.25&	42.75&	36.75&	73.50$^*$&	33.75& 71.75&	75.75&	55.00&	97.25$^*$&	74.00 & 22.00&	28.50&	10.25&	62.25$^*$&	32.75 \\
& \{$\bm{\tau}$\} & 34.25&	47.50&	39.00&	82.50$^*$&	36.00&69.00&	78.25&	55.00&	98.00$^*$&	73.75&25.00&	30.25&	11.25&	67.50$^*$&	35.50\\
& \{$\bm{\alpha}$, $\bm{\beta}$, $\bm{\tau}$\} &\textbf{48.00}&	\textbf{62.75}&	\textbf{42.50}&	\textbf{95.00}$^*$&	\textbf{47.50}&\textbf{86.25}&	\textbf{91.50}&	\textbf{61.25}&	\textbf{100.0}$^*$&	\textbf{84.75}& \textbf{33.75}&	\textbf{44.75}&	\textbf{11.75}&	\textbf{90.50}$^*$&	\textbf{43.25}\\ 
\hline
% \vspace{-2ex}
\end{tabular}
\caption{The attack success rates ($\%$) of different coefficients on LFW with adversarial meshes. $^*$ indicates white-box attacks.}
\label{tab:ablation-LFW}
% \vspace{-2ex}
\end{table*}

\section{Naturalness}

\subsection{Evaluation of AT3D-M and AT3D-P}
In this section,  we quantitatively evaluate the naturalness of meshes generated by \textbf{AT3D-M} and our method \textbf{AT3D-P}. Specifically, we calculate the discrete Gaussian curvature measure\cite{cohen2003restricted} of the original and adversarial meshes. The Gaussian curvature measure estimates the smoothness of the surface that accounts much in visual quality according to \cite{cohen2003restricted}.

\begin{definition}[The discrete Gaussian curvature measure]
Assume $P$ is the vertex set of a mesh $M$, the discrete Gaussian curvature $\Phi^{G}$ is the function that associates with every (Borel) set $B\subset\mathbb{R}^3$:
    \begin{equation}
        \Phi^{G}(B)=\sum_{p\in B\cap P}g(p),
    \end{equation}
where $g(p)$ is the angle defect of mesh $M$ at point $p$, that equals $2\pi$ minus the sum of angles between consecutive edges incident on $p$.
\end{definition}

In our experiments, the set $B$ is indeed the sphere centered at some point. Thus the discrete Gaussian curvature at point $p$ can be re-formulated as:
\begin{equation}
\begin{gathered}
\Phi^{G}\left(B(p, r)\right)=\sum_{p'\in B(p,r)\cap P}g(p'),
\end{gathered}
\end{equation}
where $r$ is the radius of the sphere. To evaluate the general smoothness of a mesh, we can denote an average Gaussian curvature measure of mesh $M$ as

\begin{equation}
\begin{aligned}
    \Phi^{G}_{M}&=\frac{1}{\left\lvert P\right\rvert}\sum_{p\in P}\lvert\Phi^{G}\left(B(p,r)\right)\rvert\\
    &=\frac{1}{\left\lvert P\right\rvert}\sum_{p \in P}\left\lvert\sum_{p'\in B(p,r)\cap P}g(p')\right\rvert.
\end{aligned}
\end{equation}

%%%%%%%%%%%%%%caption %%%%%%%%%%%%%%%%
\begin{figure}[t]
\begin{center}
\includegraphics[width=0.99\linewidth]{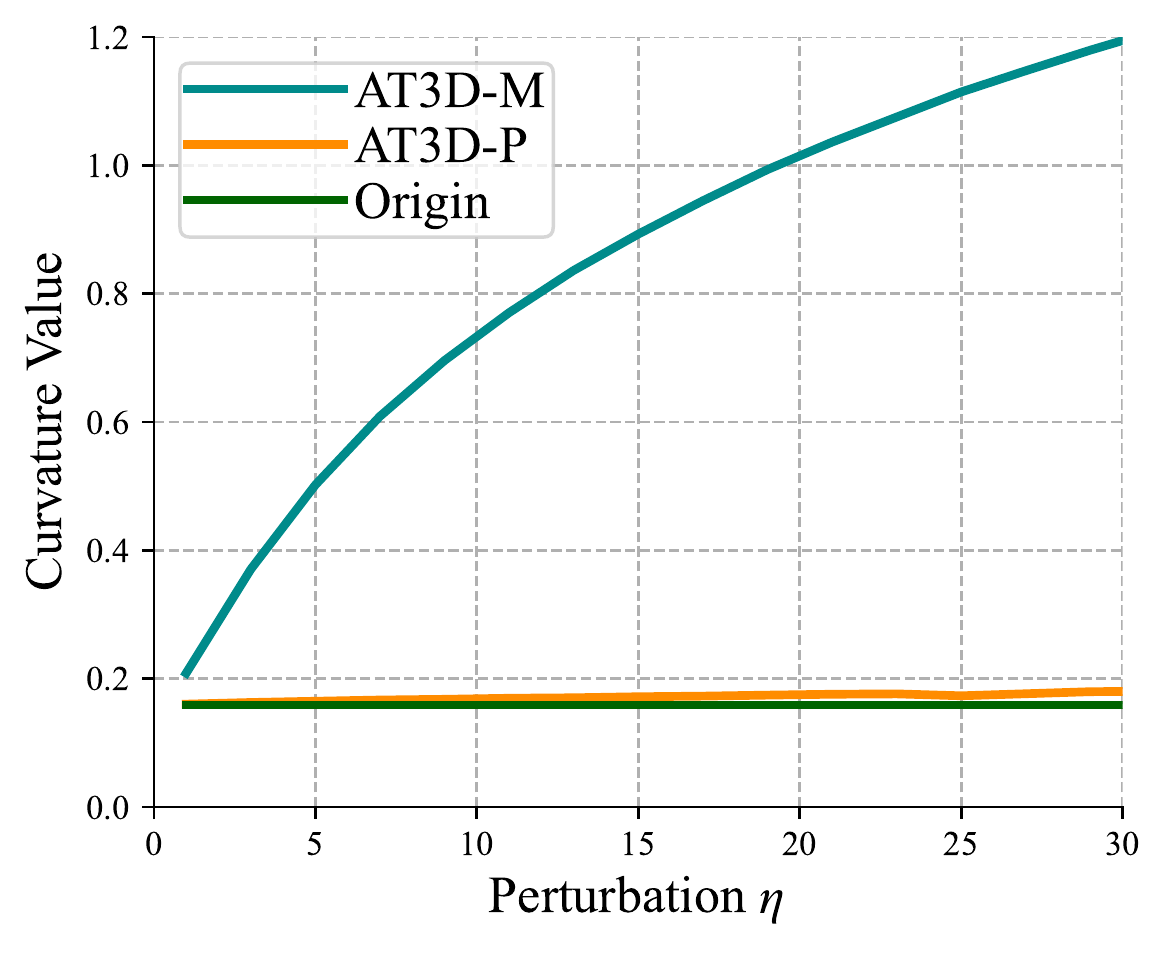}
\end{center}
\caption{Comparison of naturalness metric among different methods. }
\label{fig:nat}
\end{figure}
%%%%%%%%%%%%%%%%%%%%%%%%%%%%%%%%%%%%%

%%%%%%%%%%%%%%%%%%%% exps
\begin{figure*}[t]
\begin{center}
\includegraphics[width=0.99\linewidth]{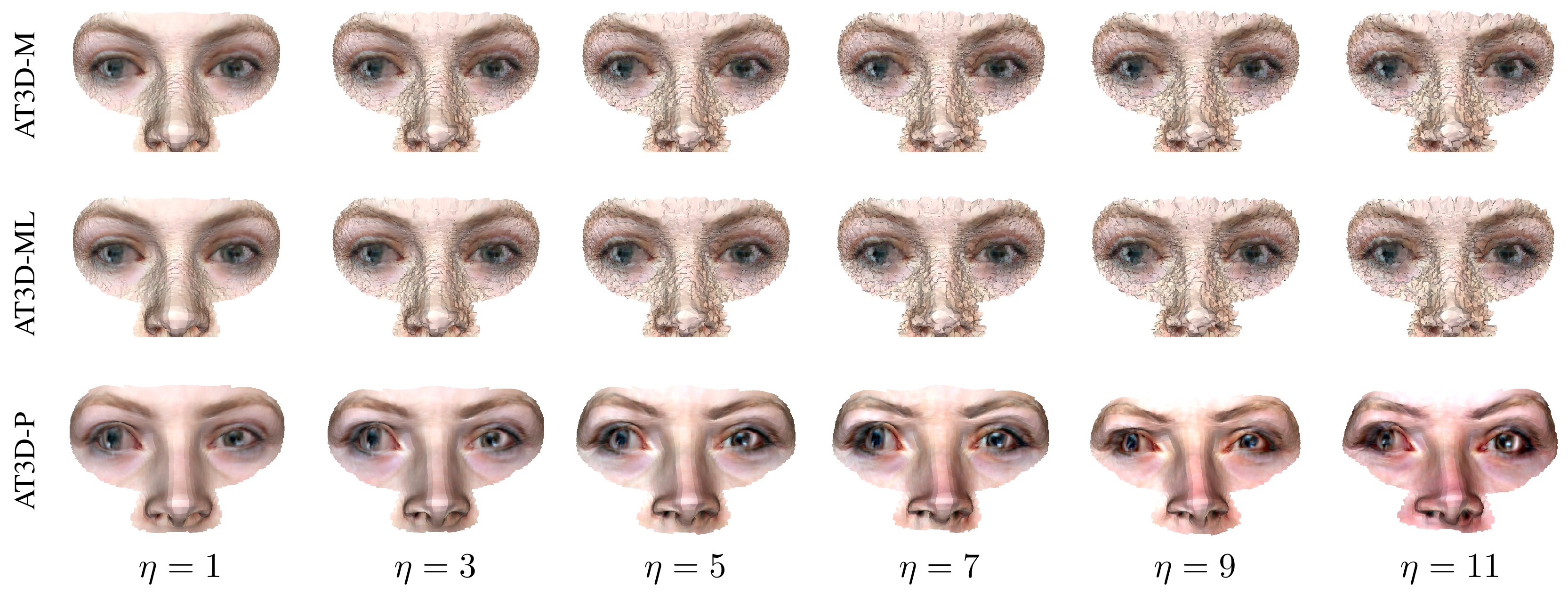}
\end{center}
% \vspace{-2ex}
\caption{Experiments on how different $\eta$ affects the performance on LFW.}
\label{fig:exs}
\end{figure*}

%%%%%%%%%%%%%%%%%%%%%%%%%%%
\begin{figure*}[t]
\begin{center}
\includegraphics[width=0.99\linewidth]{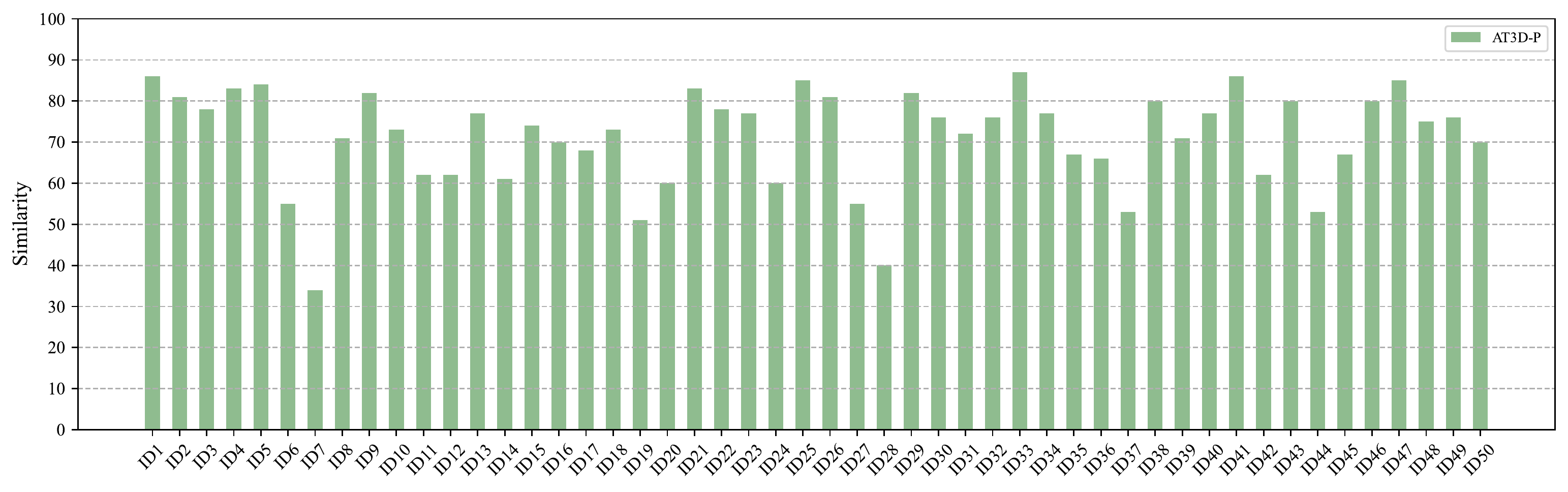}
\end{center}
% \vspace{-2ex}
\caption{Detailed physical results for every testing pair against the device {\color{blue}S-1}. }
\label{fig:pair}
\end{figure*}
%%%%%%%%%%%%%%%%%%%%%%%%%%%%

Fig.\ref{fig:nat} shows the mean value of the average Gaussian curvature measures from the original meshes, the adversarial meshes generated by \textbf{AT3D-M} and \textbf{AT3D-P} on LFW, respectively. As the perturbation $\eta$ increases, the mean curvature of meshes generated by \textbf{AT3D-M} also rapidly grows, which means the method cannot reserve the smoothness of original meshes. As a comparison, our method \textbf{AT3D-P} almost keeps the initial curvature and hardly changes the smoothness, resulting in better visual quality in terms of different values of perturbation.

\subsection{More Examples of {AT3D-ML}}
{AT3D-ML} adopted multiple popular losses in mesh-based optimization, \eg, chamfer loss, laplacian loss, and edge length loss~\cite{zhang20213d}, which are blended into the crafted AT3D to improve effectiveness and smoothness. Fig.\ref{fig:exs} shows the generated meshes of {AT3D-M}, {AT3D-ML} and {AT3D-P} under different perturbation $\eta$, respectively. As $\eta$ increases, the results of {AT3D-M} and {AT3D-ML} have rapid changes in surface curvature with severe self-intersection. We also found that {AT3D-ML} can only slightly improve the visual quality of meshes. One potential reason is that the total number of vertices and faces is too large. For example, our patch covering eyes and nose contains $35,709$ vertices and $18,368$ faces, which are very typical in 3D face models due to the requirement of high-fidelity reconstruction. However, it will make AT3D-ML difficult to keep naturalness by only applying such losses.  As a comparison, our proposed {AT3D-P} method provides better visual quality and has fewer self-intersecting faces by perturbing the 3DMM coefficients of identity and expression. Therefore, the crafted meshes can be more easily fabricated into a solid patch using 3D printers. 

% As for shape changes, compared to just perturbing the point coordinates in a larger area in \textbf{AT3D-M}, larger $\eta$ means a larger pose and greater distance between the identity features of the victim's face and those of the adversarial patch 

\section{Physical Experiments}

\textbf{3D-printed techniques.}
We choose a common and popular 3D printer, \ie, Stratasys J850 Prime, for printing all physical adversarial meshes by using resin-based materials.

\textbf{Detailed experiments.} In physical experiments, we choose \textbf{50} attacker-to-victim pairs to conduct the experiments to verify the effectiveness of the proposed method in the physical world. The procedure is evaluated by: 1) taking a face photo of a volunteer with a fixed camera under natural light; 2) crafting adversarial textured meshes for each volunteer; 3) achieving 3D printing and pasting them on real faces of the volunteers; 4) testing the attack performance against practical face recognition system. 
% We show all 3D-printed adversarial meshes in Fig.~\ref{fig:allphy}. 
For the practical device {\color{blue}S-1}, we can easily import the victims' information in batches into the system and obtain the output similarity scores when achieving attacker-to-victim adversarial testing. Therefore, we conduct the whole experiments on 50 attacker-to-victim pairs against the device {\color{blue}S-1} and present all detailed results for every testing pair, as shown in Fig.~\ref{fig:pair}. The default threshold of verification for the device {\color{blue}S-1} is $70$. If the distance of an image exceeds the threshold, the device views it as a successful verification; otherwise a failing verification. We can see that there exist $33$ successful cases among all $50$ examples. After considering anti-spoofing, we obtained a passing rate of $23/50$ as reported in this paper. Finally, we also provide video demos in the supplementary material, where 3D-printed meshes can unlock one mobile phone and an automated surveillance system.

\end{document}